\documentclass{article}

\usepackage[preprint]{corl_2026} 

\usepackage{hyperref}       
\usepackage{amsmath, amssymb, amsthm, mathtools}
\usepackage{url}            
\usepackage{booktabs} 
\usepackage{colortbl}
\usepackage{amsfonts}       
\usepackage{graphicx}
\usepackage{multirow}
\usepackage{multicol}
\usepackage{algorithm}
\usepackage{algorithmic}
\usepackage{wrapfig}
\usepackage{subfigure}
\usepackage[most]{tcolorbox}
\usepackage{graphicx}
\usepackage{booktabs} 
\usepackage{tabularx}   
\usepackage{amsfonts}  
\usepackage{amssymb}   
\usepackage[table]{xcolor} 

\definecolor{lightyellow}{RGB}{255, 247, 222}
\definecolor{reorange}{RGB}{230, 60, 0}
\definecolor{lightgray}{gray}{0.9}

\title{
  \raisebox{-.18\height}{\includegraphics[width=0.07\textwidth]{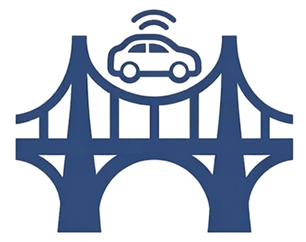}} 
  BridgeSim: Unveiling the OL-CL Gap in End-to-End Autonomous Driving
}

\author{
    Seth Z. Zhao$^1$\thanks{Equal contribution. Project leads contact: \texttt{sethzhao506@g.ucla.edu, luw015@ucsd.edu}.}, 
    Luobin Wang$^2$\footnotemark[1],
    Hongwei Ruan$^2$,
    Yuxin Bao$^1$, 
    Yilan Chen$^2$,
    Ziyang Leng$^1$,
    \AND
    Abhijit Ravichandran$^2$,
    Honglin He$^1$,
    Zewei Zhou$^1$,
    Xu Han$^1$,
    Abhishek Peri$^3$, 
    \AND
    Zhiyu Huang$^1$,
    Pranav Desai$^3$,
    Henrik Christensen$^2$,
    Jiaqi Ma$^1$, 
    Bolei Zhou$^1$\thanks{Corresponding author: \texttt{bolei@cs.ucla.edu}.} \\
    $^1$UCLA~~~~~~$^2$UCSD~~~~~~$^3$Qualcomm \\
    \tt{\href{https://vail-ucla.github.io/BridgeSim/}{\tt https://vail-ucla.github.io/BridgeSim/}}}

\begin{document}
\maketitle


\begin{abstract}
    Open-loop (OL) to closed-loop (CL) gap (\textbf{OL-CL gap}) exists when OL-pretrained policies scoring high in OL evaluations fail to transfer effectively in closed-loop (CL) deployment. In this paper, we unveil the root causes of this systemic failure and propose a practical remedy. Specifically, we demonstrate that OL policies suffer from \textit{Observational Domain Shift} and \textit{Objective Mismatch}. We show that while the former is largely recoverable with adaptation techniques, the latter creates a structural inability to model complex reactive behaviors, which forms the primary OL-CL gap. We find that a wide range of OL policies learn a biased Q-value estimator that neglects both the reactive nature of CL simulations and the temporal awareness needed to reduce compounding errors.  To this end, we propose a Test-Time Adaptation (TTA) framework that calibrates observational shift, reduces state-action biases, and enforces temporal consistency. Extensive experiments show that TTA effectively mitigates planning biases and yields superior scaling dynamics than its baseline counterparts. Furthermore, our analysis highlights the existence of blind spots in standard OL evaluation protocols that fail to capture the realities of closed-loop deployment.
    
\end{abstract}



\section{Introduction}

A growing number of works in end-to-end (E2E) autonomous driving have directed their efforts toward achieving high scores on open-loop (OL) benchmarks such as nuScenes~\cite{caesar2020nuscenes} and NAVSIM~\cite{navsimv2, navsimv1}, with some reporting performance at or near human level~\cite{RAP, DrivoR}. However, the trajectories output from OL policies frequently become infeasible or unsafe when deploying in closed-loop (CL) simulation, which shows strong discrepancy with their claimed behaviors under OL evaluations~\cite{jia2024bench, li2024ego, ol_cl_survey,zhao2025quantv2x}. We refer to such a discrepancy as the \textbf{OL-CL gap} in E2E driving.

\begin{figure}[tbh]
    \centering
    \includegraphics[width=.99\linewidth]{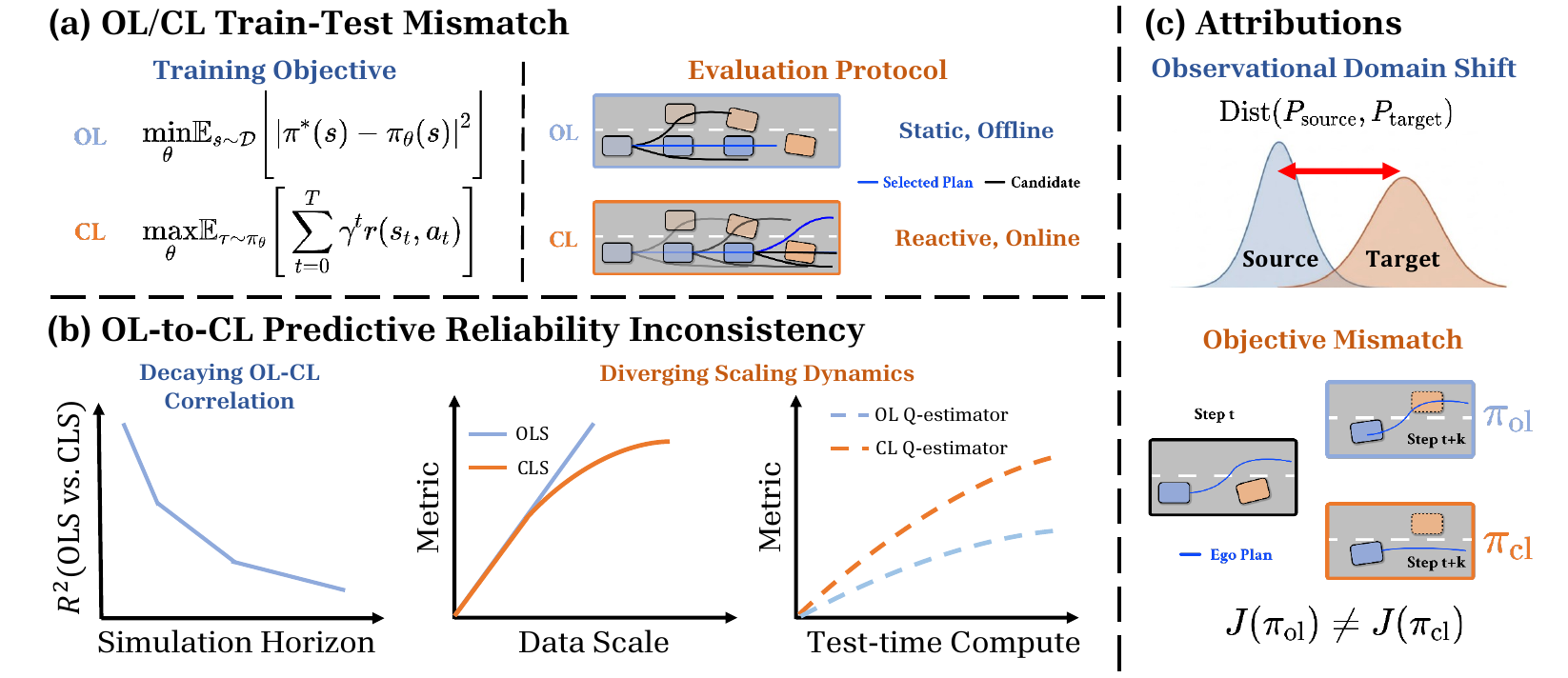}
    \vspace{-1em}
    \caption{\textbf{Motivation.} \textbf{(a)}: Open-loop (OL) train-test paradigms do not necessarily align with the objectives and evaluation metrics in closed-loop (CL) deployment; \textbf{(b)}: This mismatch results in significant performance discrepancies, characterized by decaying correlation and diverging scaling dynamics; \textbf{(c)}: We attribute this OL-CL gap to \textit{observational domain shift} and \textit{objective mismatch}.}
    \label{fig:bridgesim_hero}
    \vspace{-2em}
\end{figure}

Unveiling the OL-CL gap matters for avoiding hallucinated conclusions from OL benchmarks. We illustrate our motivations in Fig.~\ref{fig:bridgesim_hero}. First, success in OL benchmarks might create a misleading proxy for driving behavior in realistic conditions. Recent works~\cite{jia2024bench, li2024ego, ol_cl_survey} have demonstrated a lack of correspondence between OL and CL performance under similar domain settings. Furthermore, this correlation decays as the simulation horizon increases (see Fig.~\ref{fig:empirical_objective_mismatch}). Second, OL scaling laws fail to predict CL scaling. While recent studies show that OL performance scales predictably with data volume and test-time compute, neither of these relationships transfers reliably to CL deployment~\cite{zheng2024datascalinglaw, naumann2025datascalinglaw}. These discrepancies suggest that OL evaluation protocols, while being a practical and scalable proxy for costly CL simulation~\cite{caesar2020nuscenes, caesar2021nuplan, navsimv1, navsimv2, li2024pretrain}, might be biased and overlook the underlying objectives that real-world driving policies should optimize. 

In this paper, our goal is not to criticize OL evaluation protocols or the development of methods drawn from OL evaluations, but rather to clarify \textit{where} policies might fail and \textit{what} adaptations are needed to mitigate failures in CL deployment. Our key contributions are summarized as follows:

\textbf{1. Identify the fundamental failure modes of the OL-CL gap in E2E autonomous driving}. We trace the OL-CL gap to two distinct sources. First is \textit{Observational Domain Shift}, where sensor observations at test time differ from those seen during training. This distributional shift degrades perception before planning even begins (so-called sim-to-sim\footnote{Here we refer to the training and testing environments as different simulation environments: for example, NAVSIM~\cite{navsimv1,navsimv2} is a simulation environment with real-world observations, whereas Bench2Drive~\cite{jia2024bench} is a simulation environment with game-engine based observations.} gap~\cite{garcia2025road}). While our study underscores that domain adaptation can restore much of the lost perceptual fidelity, resolving the domain gap alone does \textit{not} prevent CL failure. The second and more fundamental source is the \textit{Objective Mismatch} between OL training and CL testing. More specifically, OL training optimizes against static trajectory-matching objectives that do not correspond to true closed-loop returns, and in doing so produces Q-value estimates that are biased and overfit to the pretrained simulators. Such bias is structural and persists regardless of how well the perceptual inputs are calibrated. 

\textbf{2. Propose a test-time adaptation (TTA) framework that mitigates CL deployment failure}. The framework comprises a \textit{flow-matching-based observational calibrator} that performs cross-domain observational alignment between the training and deployment environments, and a \textit{training-free adaptation mechanism} that addresses structural misalignment. Specifically, we adapt the OL policy via an \textit{truncated Q-value estimator} that corrects the scoring bias introduced by OL training objectives, and an \textit{adaptive replan mechanism} enables reactive replanning as environmental dynamics evolve. Together, these components mitigate the above-mentioned OL-to-CL deployment failures. Across diverse scenarios in CL simulation, our approach improves CL performance by up to 22\% on baseline methods and reveals more favorable scaling behavior with adaptation than baseline policies. 

\textbf{3. Introduce BridgeSim, a cross-simulator E2E closed-loop simulation platform}. To complement existing E2E benchmarks, we develop a comprehensive platform designed for the rigorous, dynamic evaluations of any E2E driving policy under interactive, closed-loop settings.

\section{Related Work}
\textbf{Driving Simulators and Benchmarks.} Driving simulators and benchmarks have long been recognized as crucial tools for training and evaluating autonomous driving systems, providing safe, efficient, and repeatable environments for developing algorithms~\cite{alghodhaifi2021autonomous, rosique2019systematic, ye2019automated, osinski2020simulation, wu2024recent}. A variety of traditional game-engine-based and high-fidelity simulators, such as GTA V~\cite{gtav}, Sim4CV~\cite{muller2018sim4cv}, AIRSIM~\cite{shah2018airsim}, and CARLA~\cite{dosovitskiy2017carla}, offer configurable environments and complex physical condition modeling. Similarly, lightweight platforms like DeepDrive~\cite{deepdrive} and DriverGym~\cite{kothari2021drivergym} provide simplified dynamics for fast experimentation, while simulators like MetaDrive~\cite{li2022metadrive} focus on modularity, scalability, and multi-agent interaction-rich scenarios. Despite their utility, these traditional platforms often suffer from domain gaps due to handcrafted assets and limited visual realism. More recent efforts have explored data-driven and generative simulation frameworks to bridge this gap. DriveArena~\cite{yang2024drivearena} synthesizes realistic traffic scenes to facilitate iterative perception-action loops. HUGSIM~\cite{zhou2024hugsim} and RAD~\cite{gao2025rad} further advance this direction by leveraging 3D Gaussian Splatting for real-time, photorealistic scene reconstruction and closed-loop simulation. Complementing these, ReconDreamer and ReconDreamer-RL~\cite{ni2025recondreamer, ni2025recondreamerrl} demonstrate a world-model-based reconstruction framework to address limitations in representing novel trajectories. However, existing generative simulation frameworks still suffer from simulation artifacts and lack of metric annotations to evaluate E2E policies (see Appendix~\ref{appendix:current_e2e_bottlenecks}).

\textbf{End-to-end Autonomous Driving.} Traditional autonomous driving systems inherit a modular design comprising perception, prediction, and planning modules, which suffer from error propagation and information loss~\cite{luo2018fast, liang2020pnpnet, sadat2020perceive, li2024womd}. Subsequently, recent research has shifted towards an integrated end-to-end (E2E) paradigm, which maps raw sensor inputs directly to control commands. UniAD~\cite{uniad} integrates modules into a unified framework and utilizes rasterized representations for information flow. VAD~\cite{vad} further optimizes the efficiency by replacing dense grids with instance-centric vectorized representation, while TransFuser~\cite{transfuser} extends inputs to multi-modalities. More recently, DiffusionDrive and DiffusionDriveV2~\cite{diffusiondrive, diffusiondrivev2} model the multi-modal distribution of actions through diffusion policy. DrivoR~\cite{DrivoR} utilizes register tokens to avoid complex intermediate representation while maintaining superior performance.
Despite the improved performance on planning evaluation metrics, recent works like AD-MLP, PDM-Close, BEV-Planner, LEAD, and RAP~\cite{zhai2023rethinking, dauner2023parting, li2024ego, LEAD, RAP} point out the limitations of open-loop (OL) benchmarks and reveal the necessity of bridging the OL-CL gap. While closed-loop (CL) training for E2E policies mitigates the gap, it presents greater complexity and computation costs~\cite{KarkusBeyondBC}. Instead, this work attributes the OL-CL gap in E2E driving to two distinct factors and formulates closing the gap via test-time adaptation.

\begin{table}[tbh]
\centering
\caption{Comparisons with other E2E open-loop and closed-loop benchmarks. MV-Consistency: multi-view temporal consistency, TF-Consistency: Traffic elements annotations consistency. PG: procedural map and behavior generation. See Appendix~\ref{appendix:bridgesim_platform} for more details of BridgeSim platform.}
\label{tab:benchmark_comparison}
\resizebox{\textwidth}{!}{ 
    \begin{tabular}{l|cccccc}
    \toprule
    \textbf{Name} & \textbf{Closed-loop} & \textbf{Long-term Simulation} & \textbf{MV-Consistency} & \textbf{TF-Consistency} & \textbf{Supported Maps} & \textbf{Supported Traffic Modes} \\
    \midrule
    NAVSIM~\cite{navsimv1, navsimv2}           & $\times$ & $\times$ & \checkmark & \checkmark & nuPlan & Log-replay \\
    Bench2Drive~\cite{jia2024bench}  & \checkmark & \checkmark & \checkmark & \checkmark & CARLA & IDM \\
    DriveArena~\cite{yang2024drivearena} & \checkmark & \checkmark & $\times$ & $\times$ & nuScenes & Log-replay \\
    HUGSIM~\cite{zhou2024hugsim}     & \checkmark & \checkmark & \checkmark & $\times$ &  nuScenes, Waymo, KITTI-360, PandaSet & Log-replay, IDM, Adversarial \\
    \midrule
    \textbf{BridgeSim (ours)}        & \checkmark & \checkmark & \checkmark & \checkmark &  CARLA, Waymo, nuScenes, nuPlan, PG & Log-replay, IDM, Adversarial \\
    \bottomrule
    \end{tabular}
}
\end{table}

\section{BridgeSim: A Unified Cross-Simulator Closed-loop Simulation Platform for E2E Driving Policies}
\label{section:bridgesim_platform}

To empirically investigate the OL-CL gap with quantifiable rigor, we introduce BridgeSim, a cross-simulator platform designed to evaluate OL pretrained policies within high-fidelity CL environments. BridgeSim designs a unified scenario protocol to incorporate diverse map scenarios (e.g., nuPlan~\cite{caesar2021nuplan}, WOMD~\cite{mei2022waymo}, and nuScenes~\cite{caesar2020nuscenes}) with heterogeneous traffic modes (e.g., log-replay, IDM~\cite{idm}, and adversarial policies~\cite{advbmt}) to stress-test E2E driving policies under a closed-loop simulation environment. Furthermore, BridgeSim offers a flexible deployment setting to simulate open-loop policy with varying execution frequencies and simulation horizons. Consequently, BridgeSim bridges the critical divide between OL benchmarks limited to short-term static prediction and existing CL benchmarks that often lack the comprehensive functionalities and annotations required for complex, reactive stress-testing. A comprehensive functional comparison between BridgeSim and other prominent E2E benchmarks is provided in Table~\ref{tab:benchmark_comparison}. 

\section{Decomposing the Open-Loop to Closed-Loop (OL-CL) Gap}
\label{section:decomposition_analysis}
In this section, we revisit the theoretical formulation of the E2E driving task and analyze the failure modes of OL-CL gap, namely \textit{observational domain shift} and \textit{objective mismatch}. Through empirical studies, we further illustrate the performance drop caused by those factors, which motivate the design of a test-time adaptation framework that recovers the
OL policies from failures.

\subsection{Preliminaries and Notations}
\label{subsec:preliminaries}
\noindent \textbf{POMDP Formalism.} We formalize the end-to-end driving task as a Partially Observable Markov Decision Process (POMDP), defined by the tuple $\langle \mathcal{S}, \mathcal{A}, \mathcal{O}, \mathcal{T}, \Omega, r, \gamma, \rho_0, T \rangle$. Here, $\mathcal{S}$ denotes the true underlying state space of the environment, $\mathcal{A}$ is the ego-vehicle's action space, and $\mathcal{O}$ is the high-dimensional observation space. At time step $t$, the system occupies state $s_t \in \mathcal{S}$. The agent receives an observation $o_t \in \mathcal{O}$ governed by the emission distribution $\Omega(o_t \mid s_t)$, executes an action $a_t \in \mathcal{A}$, and the environment evolves via $\mathcal{T}(s_{t+1} \mid s_t, a_t)$. The episode proceeds for a finite horizon of $T$ steps, where $\rho_0(s_0)$ represents the initial state distribution and $\gamma \in [0, 1)$ is the discount factor.

\noindent \textbf{End-to-End Driving Policy.} We consider end-to-end (E2E) driving policies represented as stochastic mappings $\pi^{d}_{\theta,\phi}$ parameterized by the observation domain $d\in\{\mathrm{source},\mathrm{target}\}$, observation encoder parameters $\theta$ and behavior decoder parameters $\phi$. We consider policies that predict $\mathbf{a}_t\in\mathcal{A}$\footnote{We use bold character $\mathbf{a}_t$ to denote a sequence of control $a_{t:t+H}$. $H$ is the fixed prediction sequence length that is determined at the training stage.}, which is a sequence of actions. Agents receive observations $o_t\in \mathcal{O}$ of underlying environment states $s_t\in S$ through domain-dependent emission distribution $\Omega^d(o_t\mid s_t)$. This setting allows us to isolate observational domain shifts from the behavioral counterpart. Specifically, an encoder $f_\theta: \mathcal{O} \to \mathcal{Z}$ maps raw observations to a latent scene representation $z \in \mathcal{Z}$, while the policy head $\pi_\phi(\mathbf{a}_t \mid z_t)$ generates the ego-vehicle's future trajectory. The joint distribution of the E2E policy can then be factorized into components over the action sequence, latent representation, and observation $(\mathbf{a}_t, z_t, o_t)$ conditioned on state $s_t$:
\begin{equation}
\label{eq:policy_state_latent_factor}
    \pi^{d}_{\theta,\phi}(\mathbf{a}_t,z_t,o_t \mid s_t) \triangleq \pi_{\phi}(\mathbf{a}_t\mid z_t)\,P_\theta(z_t\mid o_t)\,\Omega^d(o_t\mid s_t),
\end{equation}
where $P_\theta(z_t\mid o_t)$ denotes the encoder-induced conditional distribution. Recent works~\cite{diffusiondrive, diffusiondrivev2, RAP} usually use a deterministic encoder $z_t=f_\theta(o_t)$, we may write $P_\theta(z_t\mid o_t)=\delta(z_t-f_\theta(o_t))$.

\noindent \textbf{Closed-loop Objective.} We seek to maximize the performance of a driving policy $\pi$ via the expected closed-loop cumulative return $J(\pi) \triangleq \mathbb{E}_{s_0 \sim \rho_0, \tau \sim \pi} \left[ \sum_{t=0}^{T} \gamma^{t} r(s_t, a_t) \right]$, where the reward $r(s_t, a_t)$ evaluates the \textit{executed} control $a_t$, which typically constitutes only a partial subset of the full sequence $a_{t:t+H}$ generated by the policy at each time step.

\subsection{Observational Domain Shift}
\label{subsec:observational_shift}
\begin{figure}[tbh]
    \centering
    \includegraphics[width=.99\linewidth]{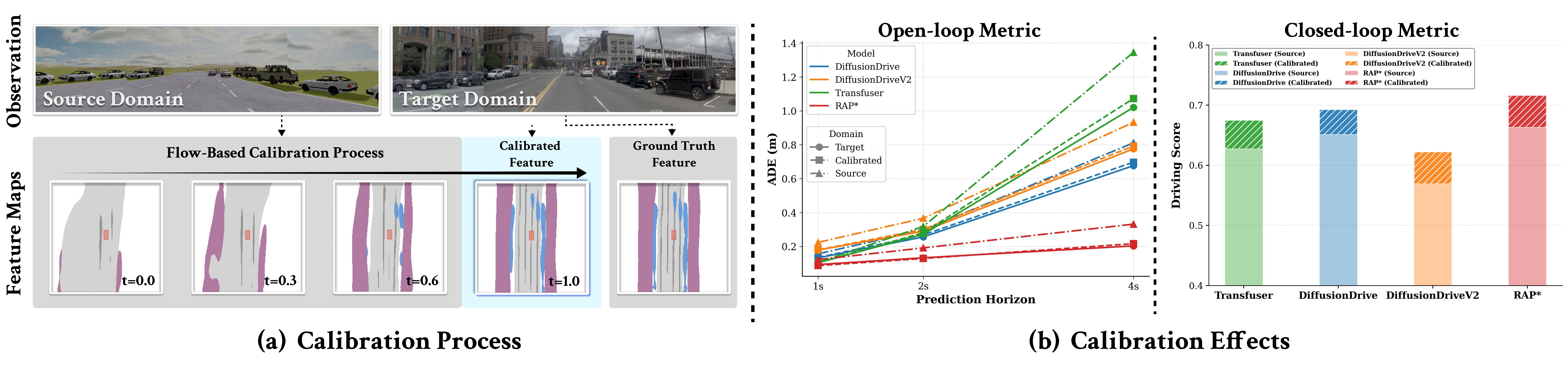}
    \vspace{-1em}
    \caption{\textbf{Empirical Study on Observational Domain Shift.} \textbf{(a)}: Flow-matching procedure for observational calibration; \textbf{(b)}: Calibration effects on OL and CL metrics. See Appendix~\ref{appendix:observational_shift_details} for experimental details.}
    \label{fig:empirical_observational_shift}
    \vspace{-2em}
\end{figure}

\textbf{Failure mode analysis.} Observational domain shift occurs when observation encoder fails to extract coherent semantic features (e.g., lane, objects, etc.) in the target environment and impairs the \textit{state observability} required for planning, as illustrated in Fig.~\ref{fig:empirical_observational_shift}(a). To denote the impact of this domain shift on task performance, we define the \emph{Observational Domain Gap} $\Delta_{\mathrm{obs}}$ as:
\begin{equation}
\small
    \Delta_{\mathrm{obs}}(\theta, \phi) \triangleq J(\pi^{\mathrm{source}}_{\theta,\phi}) - J(\pi^{\mathrm{target}}_{\theta,\phi}),
\end{equation}
where an identical control head $\pi_\phi$ isolates the degradation in expected return attributable solely to domain shift. The discrepancy between source and target domains can be measured by:
\begin{equation}
\small
    \label{eq:observation_gap}
    \mathrm{Dist}_Z(s;\theta) \triangleq \mathrm{Dist}\!\Big( P_\theta(z\mid o)\,\Omega^{\mathrm{source}}(o\mid s),\; P_\theta(z\mid o)\,\Omega^{\mathrm{target}}(o\mid s) \Big),
\end{equation}
where $\mathrm{Dist}(\cdot,\cdot)$ is the distance metric over distributions on $\mathcal{Z}$. With this formulation, we hypothesize that observational domain gap is recoverable as long as we calibrate the observation encoder $\theta$ to minimize the distributional distance.

\textbf{Empirical studies.} To test this hypothesis, we isolate and quantify $\Delta_{\mathrm{obs}}$ to examine policy behavior under observational domain shift, as depicted in Fig.~\ref{fig:empirical_observational_shift}(b). To explicitly decouple observational errors from compounding closed-loop dynamics (e.g., control drift), we conduct an open-loop ablation study. By varying only the input observation domain while holding the underlying state transitions constant, we measure the trajectory deviation from the expert ground truth following the protocol in~\cite{lin2025modelbased}. We find that under observational shift, policy performance degrades dramatically as the perception module fails to extract the semantic information necessary for planning. However, applying our practical calibration method (Section~\ref{subsec:observational_calibrator}) successfully recovers these scene features and minimizes the gap. These results unveil that: \textbf{although observational domain shift degrades perception before planning, it is largely a recoverable failure for driving policy, and targeted calibration could effectively restore its state observability and closes the domain gap.}

\subsection{Objective Mismatch}
\label{subsec:objective_mismatch}
\textbf{Failure mode analysis.} Objective mismatch occurs when OL policy's training objective optimizes for a \textit{static full-sequence objective} that ignores the \textit{reactive, receding-horizon reality} of CL deployment during test-time, where the agent predicts a trajectory of length $H$ but only executes a short prefix of $k$ steps before receiving a new observation and performing replan. In actual CL simulation, the \textbf{Unified Closed-loop Objective}\footnote{We reiterate the formulation to emphasize the importance of execution horizon $k$ in closed-loop simulation. Note that the standard closed-loop objective defined in Section~\ref{subsec:preliminaries} is a special case of Eq.~\ref{eq:executed_closed_loop_return_objective} when $k=1$.} parameterized by the execution horizon $k$ is defined as:
\begin{equation}
\small
\label{eq:executed_closed_loop_return_objective}
    J_k(\pi) = \mathbb{E}_{s_0 \sim \rho_0, \tau \sim \pi_k} \left[ \sum_{c=0}^{T/k - 1} \gamma^{c \cdot k} \sum_{i=0}^{k-1} \gamma^i r(s_{c \cdot k + i}, a_{c \cdot k + i}) \right],
\end{equation}
where $\pi_k$ denotes the policy perform replan at a frequency of $1/k$ and $c$ denotes the planning cycles. This formulation enables us to decompose the \textit{OL-to-CL} degradation brought by objective mismatch in two dimensions, namely \textbf{biased Q-value estimation} and  \textbf{compounding execution error}. 

\begin{figure}[tbh]
    \centering
    \includegraphics[width=0.99\linewidth]{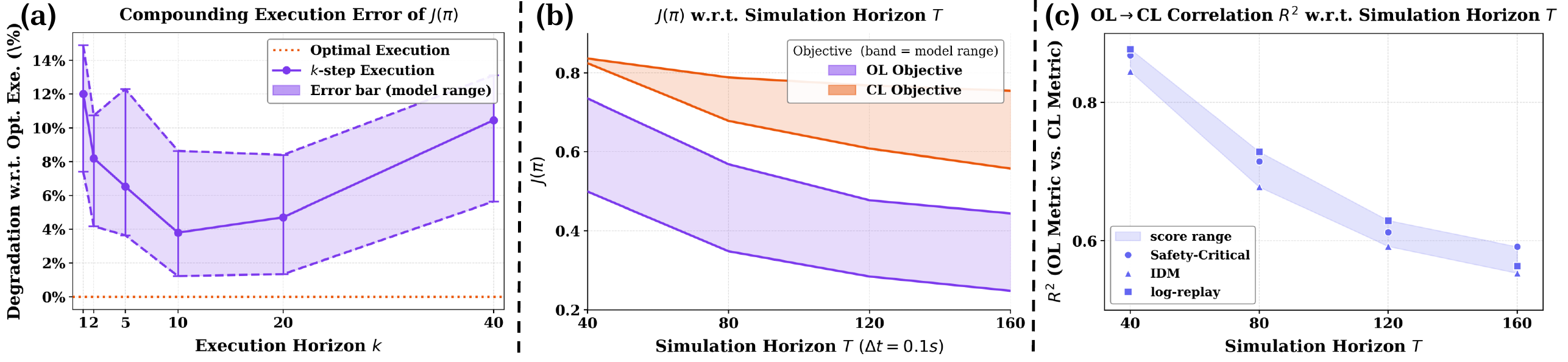}
    \vspace{-1em}
    \caption{\textbf{Empirical Study on Objective Mismatch.} \textbf{(a)}: Compounding execution error of $J(\pi)$ w.r.t. execution horizon $k$; \textbf{(b)}: $J(\pi)$ under CL and OL objectives, where the divergence between these curves constitutes the objective mismatch gap ($\Delta_{\mathrm{obj}}$); \textbf{(c)}: OL-CL Pearson correlation under different traffic modes w.r.t. simulation horizon $T$. See Appendix~\ref{appendix:objective_mismatch_details} for experimental details.}
    \label{fig:empirical_objective_mismatch}
    \vspace{-1em}
\end{figure}

Biased Q-value estimation is a byproduct of OL training, as E2E planners are typically optimized to minimize errors over the entire predicted sequence $H$. This is mathematically equivalent to optimizing an OL objective $J_{\mathrm{OL}}(\pi) \approx J_H(\pi)$, where the execution horizon $k$ is implicitly forced to equal the planning horizon $H$. Under $J_H(\pi)$, future states are unrolled via open-loop kinematic assumptions, inherently ignoring environmental stochasticity, compounding covariate shift, and the reactive behaviors of other traffic participants. Consequently, these offline proxy distributions remain fundamentally disconnected from the true closed-loop state-visitation distribution $d^{\pi_{\theta, \phi}}$ induced by high-frequency active execution.

Parallel to this, the static nature of OL training fails to instill the temporal robustness required for sequential decision-making, which makes policies susceptible to compounding execution errors. Since the policy is not trained to maintain stability across consecutive, high-frequency executions, the vanilla $k$-step execution strategy hardly achieves optimal closed-loop return as illustrated in Fig.~\ref{fig:empirical_objective_mismatch}(a). We formalize the cumulative performance degradation resulting from these factors as the \emph{Objective Mismatch Gap}:
\begin{equation}
\small
    \Delta_{\mathrm{obj}}(\theta) \triangleq J_{k^*}(\pi_{\theta, \phi_{\mathrm{CL}}}) - J_{k=H}(\pi_{\theta, \phi_{\mathrm{OL}}}),
\end{equation}
where $\phi_{\mathrm{CL}}$ represents the decoder parameters optimized under the unified CL objective (via optimal execution $k*$) over the induced distribution $d^{\pi_{\theta, \phi}}$, whereas $\phi_{\mathrm{OL}}$ represents the decoder optimized exclusively via the OL proxy ($k=H$) under the offline expert state distribution. Based on this formalization, we hypothesize that objective mismatch forms the primary structural barrier to CL transfer. Specifically, we expect that relying on the OL proxy will cause diverging optimization targets that would not maximize CL returns.

\textbf{Empirical studies.} To test this hypothesis, we empirically characterize the objective mismatch ($\Delta_{\mathrm{obj}}$) relative to the simulation horizon $T$ across a representative family of E2E policies~\cite{diffusiondrive, diffusiondrivev2, transfuser, RAP}, as illustrated in Fig.~\ref{fig:empirical_objective_mismatch}(b). As illustrated in Fig.~\ref{fig:empirical_objective_mismatch}(b), we track the divergence on the returns under OL and CL objectives. Crucially, the performance gap widens significantly exactly as $T$ exceeds $H$. This trend confirms our hypothesis, revealing that the biased Q-value estimation is insufficient for evaluating state-action values during long-horizon interaction. Furthermore, Fig.~\ref{fig:empirical_objective_mismatch}(c) shows a drastic decay in the Pearson correlation between OL and CL metrics as $T > H$, proving that standard OL evaluations might yield misleading conclusions about CL deployment readiness. Our analysis demonstrates that: \textbf{the main OL-CL gap lies in the objective mismatch, where test-time adaptation is needed to explicitly correct biased Q-value estimations and sequential decision-making during CL executions.}


\section{Proposed Method: Test-time Adaptation (TTA) Framework}
\label{section:method}

Motivated by the decomposition analysis in Section~\ref{section:decomposition_analysis}, we propose a test-time adaptation (TTA) framework that improves closed-loop robustness of OL-pretrained $\text{E2E}$ policies without additional end-to-end retraining. The framework consists of an observational calibrator that recovers latent representations via flow-matching (Section~\ref{subsec:observational_calibrator}) and a policy adaptation procedure to mitigate biased Q-value estimation and adaptively perform execution to maximize final CL return (Section~\ref{subsec:test_time_adaptation}). 

\subsection{Observational Calibrator}
\label{subsec:observational_calibrator}

As illustrated in Section~\ref{subsec:observational_shift}, we formalize the observational shift at the level of latent representations induced by the observation channel under a fixed vision encoder. Let $f_\theta:\mathcal{O}\to\mathcal{Z}$ be a pretrained encoder, and let $\Omega^{\mathrm{source}}(\cdot\mid s)$ and $\Omega^{\mathrm{target}}(\cdot\mid s)$ denote the observation channels in the source and target domains, respectively. For each domain $d\in\{\mathrm{source},\mathrm{target}\}$, given a domain-specific state distribution $\rho^{d}$, we define the induced representation distribution $\mathcal{D}_{d}$ as the marginal law of $z=f_\theta(o)$ under the generative process $s\sim\rho^{d}$ and $o\sim\Omega^{d}(\cdot\mid s)$. Note that $\mathcal{D}_{\mathrm{source}}$ and $\mathcal{D}_{\mathrm{target}}$ are both supported on $\mathcal{Z}$ but generally differ due to the domain-dependent sensing model. 

We then use flow-matching to learn a transport map $g:\mathcal{Z}\to\mathcal{Z}$ that aligns these distributions in the push-forward sense, i.e., $g_{\#}\mathcal{D}_{\mathrm{source}}\approx \mathcal{D}_{\mathrm{target}}$ \cite{lipman2022flow}, so that latent features extracted from source-domain observations can be mapped into representations that are compatible with the downstream policy $\pi_\phi$. To improve the alignment performance, we do not rely solely on a vanilla flow-matching objective; instead, we optimize the calibrator using a multi-objective loss inspired by \cite{liu2025flowing}. We show that such calibration effectively mitigates the observational domain shift by demonstrating its latent feature alignment and the resulting recovery of OL and CL metrics in Fig.~\ref{fig:empirical_observational_shift}. See Appendix~\ref{appendix:observational_calibrator} for the exact formulation of the multi-objective loss and further implementation details about dataset construction and model architecture.

\begin{figure}[tbh]
    \centering
    \includegraphics[width=0.99\linewidth]{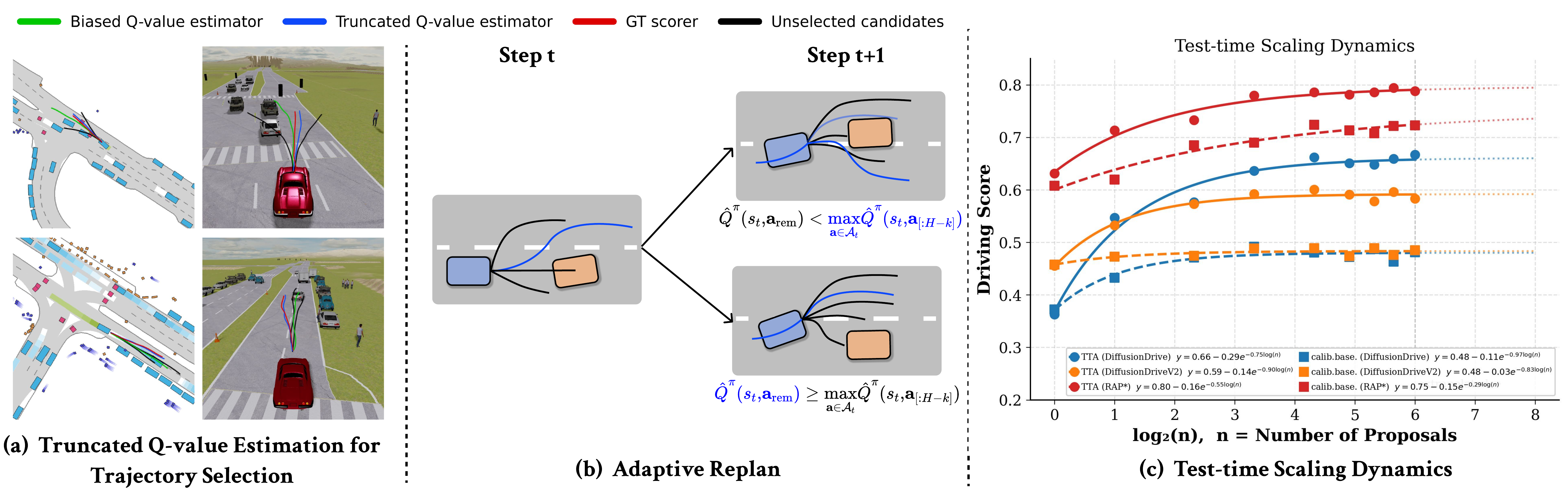}
    \vspace{-1em}
    \caption{\textbf{Test-time Policy Adaptation.} \textbf{(a)}: Truncated Q-value estimator selects safe planned trajectories during closed-loop execution; \textbf{(b)}: Adaptive replan preserves previously verified planned trajectories if the newly proposed trajectory doesn't obtain higher Q-values; \textbf{(c)}: Test-time scaling dynamics between biased (baseline) and truncated Q-value estimators (TTA).}
    \label{fig:test_time_policy_adaptation}
    \vspace{-1em}
\end{figure}

\subsection{Test-time Policy Adaptation}
\label{subsec:test_time_adaptation}

\subsubsection{Truncated Q-value Estimation}
\label{subsubsec:intrinsic_q_estimation}

To address the bias inherent in Q-values estimator learned via open-loop proxy objectives, we propose a dynamic, test-time truncated action-value estimator.
At any decision step $t$, the stochastic policy $\pi^{\Omega}_{\theta, \phi}(\cdot \mid s_t)$ outputs a discrete set of $N$ candidate trajectory-actions, denoted as $\mathcal{A} = \{\mathbf{a}^{(1)}, \dots, \mathbf{a}^{(N)}\}$. Each candidate $\mathbf{a}^{(i)}$ represents a distinct spatial-temporal plan composed of primitive controls over the full planning horizon $H$.

Evaluating the quality of a chosen prefix trajectory typically relies on the sum of immediate rewards over the $k$-step execution prefix and the expected future return $\gamma^k \mathbb{E} [ V^\pi(s_{t+k}) ]$. However, this standard expected return implicitly accumulates rewards infinitely into the future. Relying on an open-loop transition model to estimate states far beyond the explicit planning horizon $H$ inherently introduces compounding covariate shifts and unmodeled environmental stochasticity, making the unbiased estimation intractable in practice.

To overcome this, we bound the estimation strictly to the reliable planning horizon $H$ by introducing a truncated action-value estimator:
\begin{equation}
\small
    \label{eq:q_estimator_exact}
    \hat{Q}^{\pi}(s_t, \mathbf{a}_t) \triangleq R_k(s_t, \mathbf{a}_t) + \gamma^k \mathbb{E} \left[ V^\pi(s_{t+k}) \right] - \gamma^H \mathbb{E} \left[ V^\pi(s_{t+H}) \right],
\end{equation}
where $\gamma \in [0, 1)$ is the discount factor, and $R_k(s_t, \mathbf{a}_t) = \sum_{i=0}^{k-1} \gamma^{i} r(s_{t+i}, \mathbf{a}_{t+i})$ captures the immediate geometric and historical penalties strictly over the execution prefix $k$. 

By explicitly subtracting the discounted expected value at the terminal horizon limit $\gamma^H \mathbb{E} [ V^\pi(s_{t+H}) ]$, we mathematically cancel out the intractable infinite tail. This ensures that the value estimation beyond $H$ is safely ignored, restricting the estimator to evaluate only the high-confidence window bounded by the model's explicit spatial-temporal plan. The implementation of the expected Q-function is detailed in Appendix~\ref{appendix:test-time-adaptation}.

As illustrated in Fig.~\ref{fig:test_time_policy_adaptation}(a), during closed-loop execution, rather than defaulting to the best-scoring sequence from the biased open-loop policies~\cite{diffusiondrivev2, diffusiondrive}, the agent selects the candidate trajectory that maximizes $\hat{Q}^{\pi}$. Operating continuously at each decision step, this mechanism dynamically filters out plans that suffer from the biased estimation defined in Section~\ref{subsec:objective_mismatch}.

\subsubsection{Adaptive Replan}
\label{subsubsec:adaptive_replan}

Standard test-time scorers typically operate in a memoryless fashion by selecting exclusively from the current candidate set $\mathcal{A}_t$ proposed by the policy, which often leads to high-frequency action chattering and temporal inconsistency across consecutive planning cycles. To overcome compounding execution errors, we preserve the unexecuted waypoints of a previously verified plan and propose an adaptive replan mechanism that adaptively executes replan when needed, as illustrated in Fig.~\ref{fig:test_time_policy_adaptation}(b).

Let $\mathbf{a}^*_{\mathrm{prev}}$ be the optimal trajectory selected during the previous planning cycle (at step $t-k$). At the current decision step $t$, the agent has completed the $k$-step execution prefix. To rigorously evaluate whether to retain the existing plan, we extract its unexecuted remainder, containing $H-k$ waypoints, and spatially transform it into the ego-centric coordinate frame at $s_t$. We denote this transformed persistent plan as $\mathbf{a}_{\mathrm{rem}}$. To ensure a fair comparison against the newly generated candidates $\mathbf{a} \in \mathcal{A}_t$, we evaluate the truncated action-value of $\mathbf{a}_{\mathrm{rem}}$ against the corresponding $(H-k)$-step prefix of each new candidate, which we denote as $\mathbf{a}_{[:H-k]}$. The expected return is estimated using the $\hat{Q}^{\pi}$ formulation in Eq.~\ref{eq:q_estimator_exact} integrated over this shortened horizon.

We retain the persistent plan if its estimated $Q$-value equals or exceeds that of the corresponding prefix of any new candidate:
\begin{equation}
\small
    \mathbf{a}_t^* = 
    \begin{cases} 
        \mathbf{a}_{\mathrm{rem}} & \text{if } \hat{Q}^{\pi}(s_t, \mathbf{a}_{\mathrm{rem}}) \ge \max\limits_{\mathbf{a} \in \mathcal{A}_t} \hat{Q}^{\pi}(s_t, \mathbf{a}_{[:H-k]}), \\ 
        \arg\max\limits_{\mathbf{a} \in \mathcal{A}_t} \hat{Q}^{\pi}(s_t, \mathbf{a}) & \text{otherwise}.
    \end{cases}
\end{equation}
The agent abandons the persistent plan only when a newly generated candidate yields a strictly superior expected return over the shared temporal horizon.

\section{Experiments}
\label{section:experiments}

\subsection{Experimental Settings}
We utilize the BridgeSim platform as the simulation engine and benchmarks where the scenarios and traffic modes could be loaded via scenario descriptions as referred in Appendix~\ref{appendix:bridgesim_platform}. Baseline methods include DiffusionDrive~\cite{diffusiondrive}, DiffuionDriveV2~\cite{diffusiondrivev2}, and RAP~\cite{RAP}, which are pretrained in NAVSIM~\cite{navsimv2}. We evaluate the proposed TTA framework under: 1) \textbf{In-domain evaluation}: the model would be tested in the log-replay scenarios in NavHard test set under different simulation horizons, and 2) \textbf{Unseen scenes evaluation}: the model would be tested in the log-replay scenarios different from pretrained domains, e.g., nuScenes~\cite{caesar2020nuscenes}, WOMD~\cite{mei2022waymo}; 3) \textbf{Reactive evaluation}: the model would be tested in NavHard scenarios where traffic agents would be replaced with IDM~\cite{idm} or adversarial mode~\cite{advbmt}. The closed-loop metrics include a \textit{BridgeSim Driving Score} (\textbf{DS}), which is a composite score of Extended Predictive Driver Model Score (\textbf{EPDMS}) score and Route Completion (\textbf{RC}). Note that \textbf{EPDMS} is a composite metric based on No At-Fault Collisions (\textbf{NC}), Drivable Area Compliance (\textbf{DAC}), Traffic Light Compliance (\textbf{TLC}), Driving Direction Compliance (\textbf{DDC}), Lane Keeping (\textbf{LK}), Time-to-Collision (\textbf{TTC}), History Comfort (\textbf{HC}), and Extended Comfort (\textbf{EC}). See Appendix~\ref{appendix:eval_metrics} for metric details and Appendix~\ref{appendix:cl_eval_details} for more experimental details.

\subsection{Main Results}
\begin{figure}[tbh]
    \centering
    \includegraphics[width=0.99\linewidth]{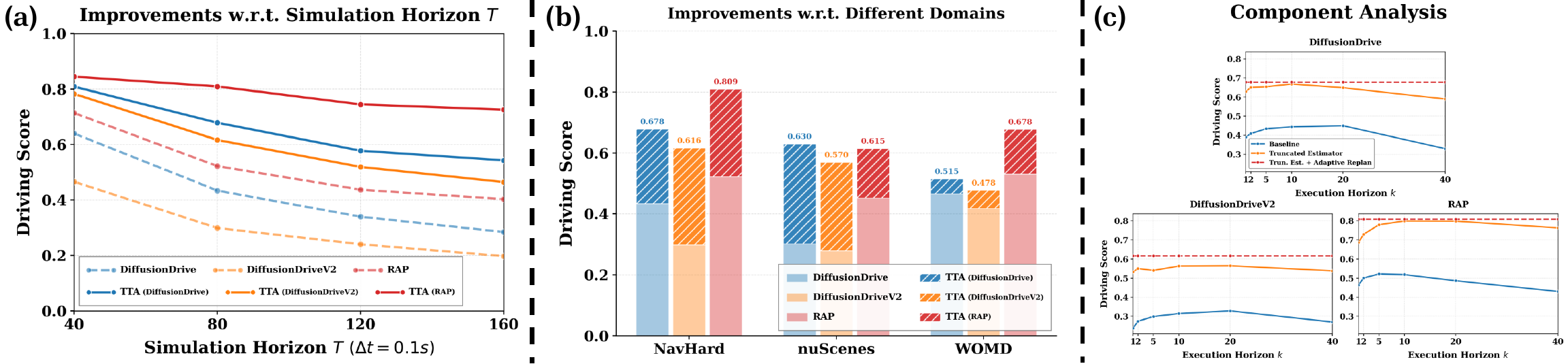}
    \vspace{-1em}
    \caption{\textbf{Improvements of proposed TTA framework.} (a): Performance gain under different simulation horizons on BridgeSim-NavHard scenarios; (b): Performance gain under different log-replay domains; (c): Component analysis of TTA framework.}
    \label{fig:main_exp}
    \vspace{-1em}
\end{figure}

\textbf{In-domain evaluation results.} As illustrated in Fig.~\ref{fig:main_exp}(a), we evaluate the log-replay scenarios under the BridgeSim-NavHard scenarios under different simulation horizons of 4/8/12/16 seconds. The reliability of E2E planners typically degrades as the simulation horizon increases due to compounding covariate shifts. Noticeably, the integration of TTA significantly flattens this decay curve across all baseline agents. By performing test-time adaptation that aligns the CL, the framework mitigates the \textit{Objective Mismatch} issue. The results demonstrate that TTA allows the agents to maintain higher temporal consistency and safety scores in long-horizon CL simulation.

\textbf{Unseen scenes evaluation results.} As illustrated in Fig.~\ref{fig:main_exp}(b), the proposed TTA framework yields consistent performance gains when integrated with diverse open-loop baselines~\cite{diffusiondrive, diffusiondrivev2,RAP}. These improvements are sustained across three distinct evaluation environments, namely NavHard~\cite{navsimv2}, nuScenes~\cite{caesar2020nuscenes}, and WOMD~\cite{mei2022waymo}, demonstrating the framework's cross-domain robustness. 

\textbf{Reactive evaluation results.} As illustrated in Table~\ref{tab:reactive_cl_scores}, we evaluate the performance gains provided by the proposed TTA framework across diverse traffic modes. Our results indicate that TTA consistently improves key CL metrics, such as progress-related metrics (\textbf{RC}) and safety-related metrics (\textbf{TTC} and \textbf{NC}), across all baseline architectures. These results demonstrate the effectiveness of the TTA framework in achieving efficiency with robust safety constraints in reactive simulations.

\textbf{Component Analysis.} As illustrated in Fig.~\ref{fig:main_exp}(c), we decouple the contributions of each TTA module to evaluate their individual impact on final performance. Our results indicate that the proposed truncated Q-value estimator significantly outperforms the original biased baseline across all tested execution horizons $k$. Furthermore, the addition of the adaptive replan mechanism consistently identifies optimal rollout trajectories and effectively mitigates compounding execution errors, leading to more robust and stable driving scores across different backbones.

\begin{table}[tbh]
\centering
\caption{Performance comparisons with different reactive traffic modes. Note that TTA improves key closed-loop metrics such as \textbf{NC}, \textbf{TTC}, and \textbf{RC}, resulting in overall higher \textbf{DS}. See Appendix~\ref{appendix:detailed_reactive_metrics} for complete sub-metric scores.}
\label{tab:reactive_cl_scores}
\renewcommand{\arraystretch}{1.5} 
\setlength{\aboverulesep}{0pt} 
\setlength{\belowrulesep}{0pt} 
\resizebox{\textwidth}{!}{ 
\begin{tabular}{l *{7}{c} @{\hspace{6pt} } *{7}{c}}
\toprule
\multirow{2}{*}{\textbf{Model}} & \multicolumn{7}{c}{\textbf{IDM}} & \multicolumn{7}{c}{\textbf{Safety-Critical}} \\
\cmidrule(lr){2-8} \cmidrule(lr){9-15}
 & \textbf{DS} & \textbf{EPDMS} & \textbf{RC} & \textbf{NC} & \textbf{DAC} & \textbf{TTC} & \textbf{LK} & \textbf{DS} & \textbf{EPDMS} & \textbf{RC} & \textbf{NC} & \textbf{DAC} & \textbf{TTC} & \textbf{LK} \\
\midrule
DiffusionDrive & 45.30 & 72.95 & 59.93 & 95.41 & 85.47 & 89.73 & 89.83 & 38.68 & 68.16 & 54.77 & 93.62 & 83.86 & 84.97 & 90.71 \\
DiffusionDriveV2 & 31.94 & 65.43 & 45.85 & 94.57 & 82.09 & 89.10 & 85.42 & 28.17 & 61.45 & 43.50 & 93.15 & 80.78 & 85.72 & 85.81 \\
RAP & 60.02 & 76.85 & 77.20 & 97.68 & 88.84 & 93.07 & 78.16 & 51.83 & 70.88 & 71.22 & 95.03 & 88.16 & 86.37 & 78.26 \\
\midrule
\rowcolor{green!15} TTA \scriptsize (DiffusionDrive) & 62.68 \scriptsize \textcolor{red}{+17.4} & 84.36 & 72.28 & 96.85 & 95.63 & 91.57 & \textbf{92.43} & 55.02 \scriptsize \textcolor{red}{+16.3} & 78.95 & 67.92 & 94.73 & 95.21 & 86.64 & \textbf{91.30} \\
\rowcolor{green!15} TTA \scriptsize (DiffusionDriveV2) & 53.96 \scriptsize \textcolor{red}{+22.0} & 77.47 & 67.89 & 96.08 & 90.53 & 89.46 & 89.98 & 47.64 \scriptsize \textcolor{red}{+19.5} & 71.66 & 65.91 & 93.54 & 89.16 & 84.23 & 90.11 \\
\rowcolor{green!15} TTA \scriptsize (RAP) & \textbf{79.07} \scriptsize \textcolor{red}{+19.1} & \textbf{90.92} & \textbf{86.21} & \textbf{98.28} & \textbf{98.62} & \textbf{95.31} & 91.52 & \textbf{65.94} \scriptsize \textcolor{red}{+14.1} & \textbf{82.01} & \textbf{78.66} & \textbf{95.30} & \textbf{96.19} & \textbf{87.71} & 90.26 \\
\bottomrule
\end{tabular}
}
\vspace{-1em}
\end{table}

\textbf{Test-time Scaling Dynamics.} We show the benefits of test-time scaling of trajectory proposals in OL metrics following~\cite{diffusiondrive, diffusiondrivev2, li2025ztrs, yao2026drivesuprim}. As illustrated in Fig.~\ref{fig:test_time_policy_adaptation}(c), the comparative experiments across different scoring mechanisms in CL simulation indicate that test-time scaling dynamics is highly dependent on the quality of the scorer. Our proposed scorers consistently benefit from scaling with broader state-visitations, whereas biased Q-value estimators could not reach similar improvements. 

\section{Conclusion}
In this work, we provide an in-depth analysis of the OL-CL gap in E2E autonomous driving and show that OL optimization does not guarantee safe or stable behavior on long-horizon CL simulation. Consequently, our analysis reveals that simply scaling OL training is insufficient for realistic deployment. Instead, targeted test-time adaptation, such as our proposed TTA framework, needs to be applied to both the observation and policy modules to obtain safe and robust driving behaviors in CL deployment. Our findings also suggest that E2E driving has reached the saturation point in OL benchmarks. At this plateau, the reliability of the predictive power of OL metrics for real-world, closed-loop capabilities remains in question. For future directions, we believe that E2E driving research should shift its focus towards more realistic CL evaluation and objective-aligned learning that directly improves driving models in a CL setting.


\clearpage


\bibliography{example}  
\newpage
\appendix

\section*{Appendix}
\section{More Related Works}
\label{appendix:more_related_works}
\textbf{E2E Driving Policy Adaptation.} Adaptation of end-to-end (E2E) driving policies has long been studied under the broader problem of covariate shift in imitation learning. Behavior cloning suffers from compounding errors when policies encounter off-distribution states, motivating interactive learning methods such as DAgger that address behavioral distribution shift through on-policy data aggregation~\cite{ross2011dagger, ross2014reinforcement}. In driving-specific settings, prior studies confirm that large-scale OL imitation learning often overfits to static expert trajectories and fails under long-horizon or reactive interactions~\cite{codevilla2019exploring,zheng2024datascalinglaw}.
Orthogonal to behavioral shift, sim-to-real transfer research highlights observational distribution shift as a critical failure mode for E2E policies~\cite{dosovitskiy2017carla}. While domain randomization and representation learning improve robustness, they do not explicitly guarantee planning-oriented fidelity across heterogeneous simulators. Recent OL benchmarks enable scalable evaluation but have been shown to correlate weakly with CL reliability, underscoring the limitations of static evaluation protocols~\cite{navsimv1, navsimv2}. In contrast to prior work that addresses perception or policy shift in isolation, our work provides in-depth decomposition of the OL–CL gap and studies their interaction under controlled simulator-induced transition changes.

\section{More Details on BridgeSim Platform}
\label{appendix:bridgesim_platform}

\subsection{Motivation}
\begin{figure}[tbh]
    \centering
    \includegraphics[width=0.99\linewidth]{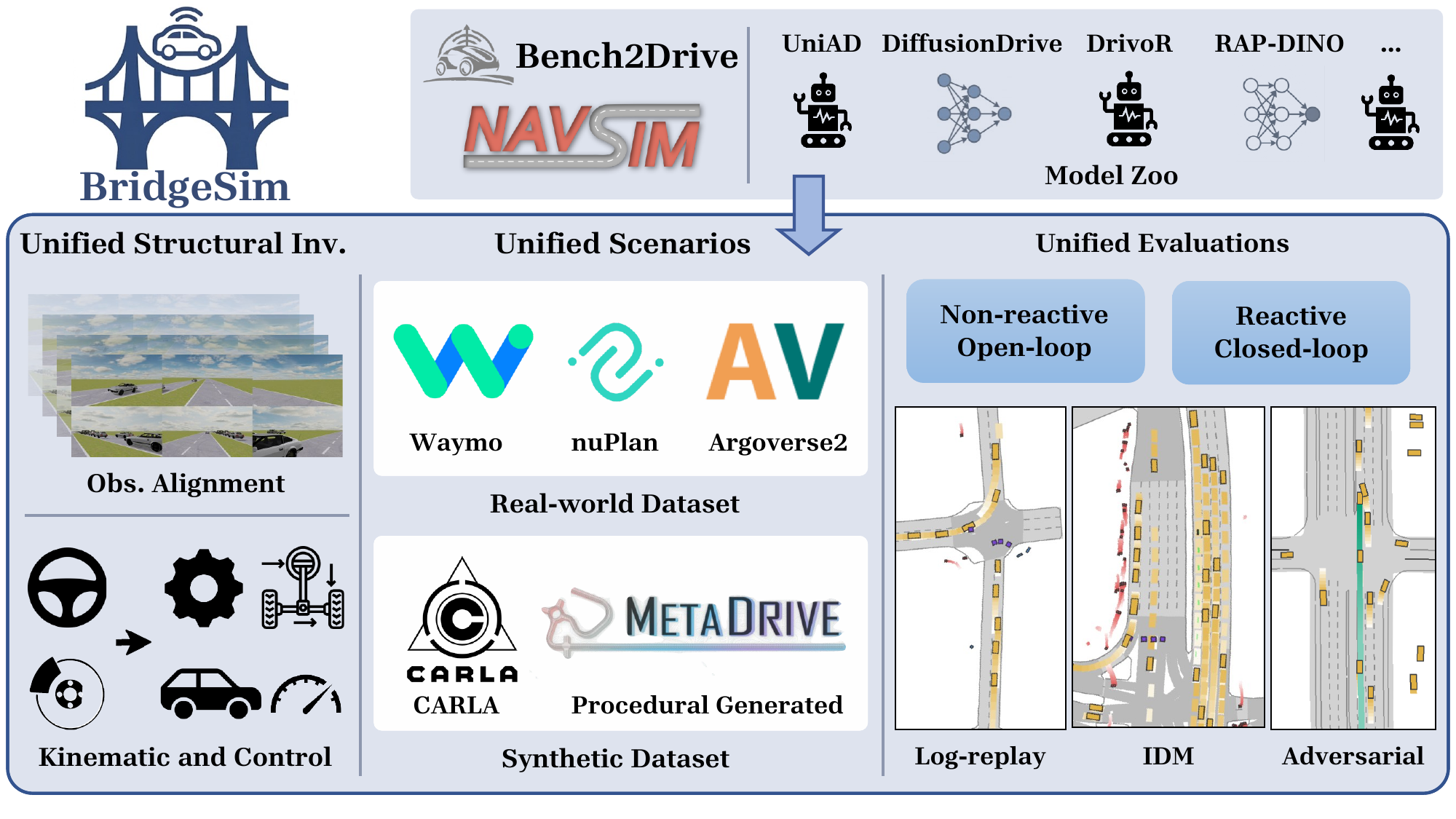}
    \vspace{-1em}
    \caption{Illustration of BridgeSim Platform. BridgeSim supports unified closed-loop simulation of policies pretrained from any other simulators, enabling us to perform rigorous decomposition analysis.}
    \label{fig:bridgesim_platform}
\end{figure}

BridgeSim platform is a unified cross-simulator closed-loop simulation platform to enable E2E policies to be evaluated across simulators, as illustrated in Fig.~\ref{fig:bridgesim_platform}. We argue that existing benchmarking testbeds are insufficient for a rigorous and unified analysis of the OL-CL gap in E2E autonomous driving. The limitations of current E2E driving benchmarks can be categorized into three primary deficiencies: 1) open-loop evaluations in existing benchmarks~\cite{navsimv1, navsimv2} often exhibit only a superficial correlation with short-term closed-loop performance and fail to serve as reliable indicators of long-horizon closed-loop capabilities; 2) current closed-loop benchmarks frequently suffer from a lack of multi-view temporal consistency~\cite{yang2024drivearena}, inaccurate annotations for observationally consistent traffic elements~\cite{zhou2024hugsim}, or a lack of environmental diversity and agent behaviors necessary to stress-test generalizability~\cite{jia2024bench} (see Appendix~\ref{appendix:current_e2e_bottlenecks} for more details); and 3) policies transferred between simulators often encounter inconsistent vehicle dynamics and controller configurations, inducing artificial performance drops that obscure the true OL-CL gap.

These systematic deficiencies necessitate the design of the BridgeSim platform, which is engineered to be controllable for identifying modular changes and consistent across evaluation modes and metrics. To this end, BridgeSim leverages the MetaDrive~\cite{li2022metadrive, li2023scenarionet} simulation engine to resolve the controllability and inconsistency issues prevalent in previous benchmarks. As illustrated in Fig.~\ref{fig:bridgesim_platform}, BridgeSim ensures high-fidelity closed-loop simulation through a modularized architecture implementing: 1) a unified structural framework for policy transfer to maintain state-action compatibility across domains; 2) unified scenario protocols that enable precise, repeatable environment replay and controllable agent interactions; and 3) unified evaluation modes and metrics to ensure fair and accurate comparisons.

\subsection{Unified Structural Invariance in Policy Transfer} 
To quantify the $\text{OL-CL}$ gap, it is imperative to decouple the policy's decision-making logic from confounding systemic variables. To establish an equitable diagnostic framework, BridgeSim enforces two levels of structural invariance:

\textit{1) Observational Alignment:} We standardize sensor configurations, including camera extrinsics and intrinsics, across all evaluated benchmarks. This standardization ensures that performance failures are strictly attributable to the policy’s representational robustness rather than superficial geometric mismatches in the input space $\Omega$. Since the simulator maintains rigorous physical properties, we further enforce multi-view temporal consistency, addressing a common deficiency in existing benchmarks. Sensor observations $o_t$ are synthesized via geometric projection, ensuring that $o_t$ remains spatially and temporally consistent with the underlying state $s_t$ across all camera frames and timesteps. This approach eliminates spurious distribution shifts, thereby ensuring that policy failures are rooted in decision-making errors rather than simulation artifacts. Representative failure cases of existing generative and Gaussian Splatting-based simulation paradigms are detailed in the Appendix~\ref{appendix:current_e2e_bottlenecks}.

\textit{2) Kinematic and Control Alignment:} We enforce consistency across vehicle dynamics and control execution layers to isolate the sources of performance degradation. By standardizing the transition dynamics $\mathcal{P}$ during the execution of predicted waypoints, we ensure that the measured $\text{OL-CL}$ gap reflects fundamental planning-level failures rather than downstream actuation discrepancies. To ensure our evaluation reflects standard practice, we adopt controllers that are ubiquitous across diverse real-world autonomous driving platforms~\cite{christensen2021autonomous}; specifically, we utilize a standard PID formulation for longitudinal control and a Pure Pursuit~\cite{coulter1992implementation} controller for lateral tracking. Furthermore, BridgeSim maintains strict structural invariance by explicitly decoupling these vehicle dynamics from the rendering pipeline.

\subsection{Unified Scenario Protocols} 
\textbf{Scenario Descriptions.} We define a unified, serialized data schema designed to encapsulate all static and dynamic elements necessary to reconstruct any driving episode. To ensure cross-platform compatibility, the converters map diverse source data from real-world logs (e.g., Waymo~\cite{mei2022waymo}, nuScenes~\cite{caesar2020nuscenes}) to synthetic environments (e.g., CARLA~\cite{jia2024bench}) into a standardized \texttt{ScenarioDescription} format.

This centralized data structure contains three primary components:

\textit{1) Map features.} BridgeSim ingests diverse real-world datasets, including Waymo~\cite{mei2022waymo}, nuPlan~\cite{caesar2021nuplan}, nuScenes~\cite{caesar2020nuscenes}, via a standardized protocol to extract map geometry. This geometry is discretized and stored as dense point sequences, each assigned specific type attributes. We subsequently vectorize HD map elements, such as lanes, boundaries, crosswalks, and stop lines, into a unified polyline representation. By ensuring the precise spatial alignment of these layouts within the simulation environment, we eliminate rendering artifacts that might otherwise distort lane properties, maintaining higher fidelity and consistency than existing benchmarks.

\textit{2) Object tracks.} Dynamic agents are stored as temporal tracks containing state sequences, including $position$, $heading$, $velocity$, and $dimensions$ alongside validity masks. To address the varying sampling rates across different datasets, we implement an interpolation pipeline that up-samples trajectories. This process utilizes linear interpolation for position and velocity, while applying angular interpolation for heading to ensure smooth kinematic playback during simulation.

\textit{3) Dynamic map states.} Traffic light states are decoupled from the static map geometry. We map individual traffic signals to specific lane IDs and store their state sequences (e.g., \textit{STOP}, \textit{WAIT}, \textit{GO}). This mapping enables the simulator to dynamically update the lane signaling during replay, ensuring that the behavioral constraints of the agents remain in sync with the visual environment.

\textbf{Traffic Modes.} BridgeSim provides an API for the dynamic configuration of agent behaviors to stress-test policies under varying levels of interactivity. We allow for the following agent behavior modes:

\textit{1) Log-replay mode.} In this mode, background agents strictly adhere to their recorded trajectories from the source dataset. The simulator essentially replays the preset vehicle tracks. While this preserves the exact real-world context, these agents remain non-reactive to the ego-vehicle's deviations.

\textit{2) IDM mode}~\cite{idm}. To bridge the gap between static datasets and closed-loop interaction, we replace log-replay policies with an Intelligent Driver Model (IDM). While background vehicles are initialized at their ground-truth poses, their subsequent behaviors are governed by heuristic policies. These agents adhere to lane geometry, maintain safety gaps, and actively respond to the ego-vehicle's presence.

\textit{3) Adversarial mode.} We provide an interface for injecting adversarial agents~\cite{advbmt} into log-replay scenarios. This enables the generation of safety-critical edge cases, challenging ego-policies to react instantaneously to avoid imminent collisions.

\textbf{Coordinate Systems.} A critical challenge in evaluating policies across different simulators is the misalignment of coordinate systems (e.g., global vs. local, left-handed vs. right-handed). BridgeSim enforces a unified right-handed coordinate system (RHS) centered on a local scenario anchor. We employ the following techniques to ensure coordinate system alignment:

\textit{1) Chirality normalization.} For datasets using left-handed systems (e.g., CARLA, Bench2Drive), we apply a basis transformation during conversion. Orientation angles (yaw, roll) are inverted accordingly to maintain kinematic consistency.

\textit{2) Spatial anchoring.} To mitigate floating-point errors inherent in large-scale global coordinates (common in nuScenes and nuPlan), we normalize all spatial data relative to a scenario anchor, defined as the ego-vehicle's position at $t=0$. All map features and trajectories are translated so that the map origin aligns with this anchor, ensuring numerical stability for the planner and simulator physics engine.

\textbf{Time Lengths.} BridgeSim decouples the simulation horizon from the source simulators by supporting long-term simulation durations determined by the full extent of the source log. The scenario duration can also be manually configured to a variable length comprising a sub-duration of the full source scenario. To accommodate the heterogeneous sampling rates of these diverse datasets, we implement a standardized trajectory interpolation protocol. This pipeline resamples source signals to the simulator’s native 10Hz control frequency using linear and angular interpolation. This ensures consistent kinematic evaluation across varying time steps without artificially truncating the scenario.

\subsection{Unified Evaluation Modes and Metrics}
\label{appendix:eval_metrics}
\textbf{Evaluation modes.} To rigorously assess the capability of driving agents to generalize from offline datasets to online deployment, BridgeSim implements a unified evaluation pipeline. This pipeline supports both \textit{non-reactive open-loop evaluation} and \textit{reactive closed-loop evaluation}, within a shared environment $\mathcal{E}$. Let $\mathcal{S}$ denote the state space and $\mathcal{A}$ the action space. At time $t$, the agent observes state $s_t \in \mathcal{S}$ and generates a planned trajectory $\tau_t = \pi(s_t)$.

\textit{1) Non-reactive open-loop simulation. }  In open-loop mode, the simulation decouples the ego-agent's trajectory planning from the environment's state transitions. The environment dynamics are strictly governed by the ground-truth trajectories from the source dataset. In this mode, the ego-vehicle's state transition is forced to match the logged trajectory $\mathcal{T}_{log}$ independent of the policy output. Specifically, the state transition is defined as $s_{t+1} = f(s_t, a^*_{t})$, where $a^*_t$ denotes the expert action recorded in the log. The model predicts a trajectory $\hat{\tau}_t$ at every timestep, which is transformed from the ego-frame to the global simulation frame for evaluation against the static map and dynamic agents (replayed from logs). This mode isolates the planner's reasoning ability without the confounding factor of cumulative control error.

\textit{2) Reactive closed-loop simulation. } In closed-loop mode, the evaluation focuses on the agent's ability to robustly control the vehicle over long horizons. The simulation environment is configurable to support either a \textit{fully reactive} paradigm, where background traffic behavior is governed by Intelligent Driver Models (IDM) that dynamically respond to the ego-agent's movements, or a \textit{semi-reactive} paradigm, where background traffic strictly adheres to the original logged trajectories while the ego-vehicle remains physically controlled by the policy.

\textbf{Evaluation metrics.} To quantify driving performance within the BridgeSim environment, we implement a set of open-loop and closed-loop metrics following previous protocols~\cite{navsimv1, navsimv2, zhou2024hugsim, yang2024drivearena, li2025hydra-mdp++}. For non-reactive open-loop simulations, we adopt the EPDMS framework~\cite{navsimv1, navsimv2, li2025hydra-mdp++}. For reactive closed-loop simulations, we incorporate a composite metric inspired by the HUGSIM~\cite{zhou2024hugsim} and DriveArena~\cite{yang2024drivearena} Driving Score. 

\textit{1) Open-loop Metric: Extended Predictive Driver Model Score} (\textbf{EPDMS}). The \textbf{EPDMS} score is implemented as a weighted combination of safety-critical constraints and driving quality features, as shown in Table~\ref{tab:epdms_metrics}. Unlike L2 displacement errors, \textbf{EPDMS} penalizes physically infeasible or unsafe plans even if they mimic the log. At every frame, we transform the model’s predicted trajectory into the simulation frame and score it against the static and dynamic map features. The final score $S_{EPDMS} \in [0, 1]$ for a given frame is defined as the product of critical boolean constraints $\mathcal{C}$ and a weighted sum of soft metrics $\mathcal{F}$:
\begin{equation}
    S_{EPDMS} = \left( \prod_{c \in \mathcal{C}} \mathbb{I}_c \right) \cdot \left( \frac{\sum_{f \in \mathcal{F}} w_f \cdot v_f}{\sum_{f \in \mathcal{F}} w_f} \right),
\end{equation}
where $\mathbb{I}_c \in \{0, 1\}$ is the indicator function for constraint compliance, $v_f \in [0, 1]$ is the normalized feature score, and $w_f$ is the corresponding weight.  

\begin{table}[tbh]
\centering
\caption{Descriptions of subscores of \textbf{EPDMS}.}
\label{tab:epdms_metrics}
\small
\renewcommand{\arraystretch}{1.0} 
\begin{tabularx}{\columnwidth}{@{} l >{\raggedright\arraybackslash}X @{}}
\toprule
\textbf{Metric} & \textbf{Description} \\ \midrule
\multicolumn{2}{@{}l}{\textbf{\textit{Critical Constraints ($\mathcal{C}$)}}} \\ \addlinespace
No At-Fault Collisions (\textbf{NC}) & Ego polygon $\mathcal{P}_{ego}$ must not intersect any replayed object polygons $\mathcal{P}_{obj}$. \\
Drivable Area Compliance (\textbf{DAC}) & Entire horizon must satisfy $\mathcal{P}_{ego} \subset \mathcal{M}_{drivable}$ within the HD map. \\
Traffic Light Compliance (\textbf{TLC}) & Agent cannot cross stop lines associated with red light states. \\
Driving Direction Compliance (\textbf{DDC}) & Heading deviation $|\theta_{ego} - \theta_{lane}|$ must remain within $\pm \pi/2$. \\ \midrule
\multicolumn{2}{@{}l}{\textbf{\textit{Quality Features ($\mathcal{F}$)}}} \\ \addlinespace
Ego Progress (\textbf{EP}) & Represents the agent progress as a ratio to an approximated safe upper bound. \\
Lane Keeping (\textbf{LK}) & Checks the ego-vehicle follows the centerline of its current lane and avoids lingering between adjacent lanes. \\
Time-to-Collision (\textbf{TTC}) & Penalizes trajectory $\hat{\tau}$ if the projected time-to-collision is too low. \\
History Comfort (\textbf{HC}) & Evaluates the planner’s predicted trajectory with a short segment of historical motion from the human driver.  \\ 
Extended Comfort (\textbf{EC}) & Limits kinematic derivatives for passenger comfort: acceleration $a \le 4.89$, jerk $j \le 8.37$, and yaw rate $\dot{\theta} \le 0.95$. \\
\bottomrule
\end{tabularx}
\end{table}

\textit{2) Closed-loop Metric: BridgeSim Driving Score.} For reactive closed-loop simulation, we define the final driving score $S_{CLS}$ as the product of the agent's global Route Completion (\textbf{RC}) and the temporal average of the frame-level scores ($S_{EPDMS}^{(t)}$):
\begin{equation}
S_{CLS} = R_c \times \frac{1}{T} \sum_{t=1}^{T} S_{EPDMS}^{(t)},
\end{equation}

where $R_c\in[0, 1]$ represents the percentage of the route completed by the agent relative to the expert driver's path or the goal destination, $T$ is the total number of frames in the simulation episode, and $S_{EPDMS}^{(t)}$ is the EPDM score at time step $t$. Note that the frame-level ego progress score (\textbf{EP}) from open-loop EPDM scoring is removed, as progress is now explicitly measured by the distance traveled to the goal in the simulator~\cite{zhou2024hugsim}. Crucially, for the closed-loop setting, the set of soft features $\mathcal{F}$ is modified to exclude displacement error or progress-based terms, focusing solely on driving quality and safety constraints. This formulation ensures that an agent achieves a high score only if it completes the route while consistently maintaining safe and comfortable driving behavior throughout the episode.

\subsection{Sim-to-Sim Evaluation Results across Simulators}

\subsubsection{Supported Model Zoo}
We include a diverse set of publicly available end-to-end driving models for zero-shot cross-simulator evaluation. Unless otherwise specified, we use the official checkpoints released by the original authors or benchmark maintainers. Note that new models could be easily incorporated into BridgeSim following the instructions in Appendix~\ref{appendix:extensibility}.

\textbf{UniAD}~\cite{uniad} is a unified end-to-end autonomous driving framework that jointly models perception, prediction, and planning through a query-based transformer architecture. We use its official Bench2Drive checkpoint in our evaluation.

\textbf{Vectorised Autonomous Driving (VAD)}~\cite{vad} is a vectorized end-to-end driving method that represents map elements and traffic agents with structured queries rather than dense raster features. We use its official Bench2Drive checkpoint.

\textbf{TransFuser}~\cite{transfuser} is a multi-modal end-to-end driving model that fuses RGB camera and LiDAR observations for waypoint prediction. We use the released NAVSIM-trained checkpoint in our evaluation.

\textbf{DiffusionDrive}~\cite{diffusiondrive} formulates autonomous driving as a generative planning problem using a truncated diffusion process with anchor trajectories. The released checkpoint used in our evaluation is trained on NAVSIM.

\textbf{DiffusionDriveV2}~\cite{diffusiondrivev2} extends DiffusionDrive by incorporating reinforcement learning into the trajectory denoising process. We use the released NAVSIM-trained checkpoint.

\textbf{LEAD}~\cite{LEAD} studies learner--expert asymmetry in imitation learning and builds on a TransFuser-v6 driving architecture. The released checkpoints include CARLA-based models for Leaderboard 2.0 / Bench2Drive-style evaluation as well as a NAVSIM checkpoint under the LTFv6 variant.

\textbf{DrivoR}~\cite{DrivoR} is a transformer-based end-to-end planner built on a pretrained DINOv2 visual encoder with token compression and trajectory scoring modules. The released model used in our evaluation is trained on NAVSIM.

\textbf{Rasterization Augmented Planning (RAP)}~\cite{RAP} is a rasterisation-based augmentation framework for end-to-end driving that improves training with counterfactual rasterised views and raster-to-real feature alignment. The official checkpoint used in our comparison is trained on NAVSIM.

\textbf{Alpamayo-R1}~\cite{wang2025alpamayo} is a vision-language-action (VLA) model for end-to-end driving model that emphasizes robust temporal reasoning to handle complex urban maneuvers. We utilize the official released checkpoint in our evaluation.

\textbf{OpenPilot}~\cite{openpilot2026} is a vision-based driving system that utilizes a multi-task convolutional neural network to predict the path, lead vehicle properties, and lane boundaries. We use the released v0.11.0 version checkpoint provided by comma.ai.

\begin{table*}[h]
\centering
\caption{Quantitative comparison of E2E driving policies evaluated on BridgeSim-NavHard. All results are reported with a replan rate of 5 and a simulation horizon of 8 seconds. We use $\mathrm{rr}=5$ to align with the temporal discretization commonly used for training on NAVSIM/OpenScene, where the dataset is provided at 2\,Hz (i.e., $\Delta t = 0.5\,\mathrm{s}$).}
\label{tab:appendix_main_results}
\resizebox{\textwidth}{!}{
\begin{tabular}{l|c|c|cc|cccccccc}
\toprule
\textbf{Model} & \textbf{Pretrained Sim} & \textbf{DS} & \textbf{EPDMS} & \textbf{RC} & \textbf{NC} & \textbf{DAC} & \textbf{DDC} & \textbf{TLC} & \textbf{TTC} & \textbf{LK} & \textbf{HC} & \textbf{EC} \\ 
\midrule

Human Expert & N/A & 91.14 & 91.96 & 98.86 & 98.18 & 99.23 & 99.76 & 98.62 & 94.55 & 90.22 & 99.10 & 71.44 \\

\midrule

UniAD {\scriptsize (CVPR'23)} & Bench2Drive & 31.00 & 57.55 & 55.25 & 93.15 & 71.75 & 99.76 & 99.61 & 81.13 & 97.70 & 74.25 & 30.04 \\
VAD {\scriptsize (ICCV'23)} & Bench2Drive & 26.25 & 57.60 & 46.76 & 93.29 & 71.22 & 99.76 & 99.68 & 82.31 & 98.67 & 71.13 & 27.67 \\

\midrule

DiffusionDrive {\scriptsize (CVPR'25)} & NAVSIM & 43.29 & 69.70 & 61.30 & 95.38 & 81.48 & 99.76 & 98.81 & 87.47 & 91.57 & 91.64 & 39.02 \\
DiffusionDriveV2 {\scriptsize (arXiv)} & NAVSIM & 29.81 & 62.01 & 45.85 & 93.48 & 79.75 & 99.76 & 98.83 & 86.93 & 84.11 & 71.38 & 20.62 \\
RAP {\scriptsize (ICLR'26)} & NAVSIM & 52.08 & 72.41 & 70.74 & 95.63 & 85.24 & 99.76 & 99.18 & 88.14 & 84.69 & 92.84 & 48.91 \\
LEAD {\scriptsize (CVPR'26)} & NAVSIM & 26.34 & 54.02 & 47.42 & 93.29 & 68.44 & 99.53 & 99.89 & 84.70 & 96.38 & 91.62 & 36.59 \\
TransFuser {\scriptsize (PAMI'23)} & NAVSIM & 41.46 & 68.17 & 61.28 & 95.73 & 78.99 & 99.76 & 98.63 & 88.79 & 89.25 & 95.06 & 49.25 \\
LTF {\scriptsize (NeurIPS'24)} & NAVSIM & 50.10 & 69.29 & 70.64 & 95.49 & 78.80 & 99.53 & 98.51 & 88.61 & 90.28 & 95.02 & 58.27 \\
DrivoR {\scriptsize (CVPR'26)} & NAVSIM & 42.79 & 66.54 & 64.66 & 95.41 & 76.34 & 100.00 & 99.46 & 95.41 & 94.37 & 96.99 & \textbf{66.69} \\

\midrule

OpenPilot & Comma.ai & 64.75 & 79.50 & 82.21 & 97.55 & 94.29 & 100.00 & 98.89 & 92.58 & 84.43 & 92.12 & 62.06 \\
Alpamayo-R1 & Nvidia & 37.78 & 64.69 & 59.65 & 94.98 & 76.54 & 100.00 & 99.73 & 87.39 & 93.63 & 94.82 & 68.24 \\

\midrule
\rowcolor{green!15} TTA \scriptsize (DiffusionDrive) & NAVSIM & 67.84 & 79.77 & 82.54 & 96.48 & 87.03 & \textbf{100.00} & 99.01 & 96.07 & 91.85 & 36.18 & 33.72 \\
\rowcolor{green!15} TTA \scriptsize (DiffusionDriveV2) & NAVSIM & 61.46 & 76.86 & 77.86 & 95.83 & 86.80 & \textbf{100.00} & \textbf{99.17} & 92.93 & 91.51 & 26.22 & 24.32 \\
\rowcolor{green!15} TTA \scriptsize (RAP) & NAVSIM & \textbf{80.92} & \textbf{91.68} & \textbf{87.06} & \textbf{96.95} & \textbf{90.14} & \textbf{100.00} & 98.90 & \textbf{98.07} & \textbf{94.72} & 46.18 & 54.60 \\

\bottomrule
\end{tabular}}
\end{table*}

\subsubsection{Sim-to-Sim Evaluation Results}
Our evaluation in Table~\ref{tab:appendix_main_results} reveals that most open-source models suffer from the OL-to-CL deployment gap when integrated into the BridgeSim platform. We observe that models pretrained on synthetic simulation domains frequently encounter simulation overfitting across both visual and behavioral aspects. Specifically, models trained on Bench2Drive exhibit a substantial performance degradation. This disparity underscores a significant sim-to-sim gap between CARLA’s synthetic rendering and BridgeSim’s real-world replayed scenarios. On the other hand, NAVSIM-pretrained models demonstrate better performance, as their training environment relies on real-world traffic behaviors. These findings emphasize that, besides visual fidelity, behavioral realism is also critical when training autonomous agents for robust simulation performance. The results show that strong in-domain performance doesn't always translate to broader generalizability.

\subsection{Extensible Design for Incorporating New Maps, Models, and Traffic Modes}
\label{appendix:extensibility}

BridgeSim possesses high backward-compatibility and extensibility to incorporate new maps, models, and traffic modes, encouraging open-source contributors to independently augment the benchmark with novel driving environments,
state-of-the-art planning models, and alternative traffic behaviors without
modifying any shared evaluation logic.

\textbf{Incorporating New Maps.} To integrate a new map, contributors only need to implement the \texttt{converter} module and integrate the new maps into the scenario description protocol. This module provides dedicated, dataset-specific pipelines for each supported source. Each pipeline implements a conversion function that transforms raw, dataset-native records into a standardized scenario description dictionary. By invoking the shared serialization utility within this function, the resulting scenario directory becomes automatically discoverable by the environment manager during evaluation, requiring no modifications to downstream evaluation logic.

\textbf{Incorporating New Models.}
Contributor could use the \texttt{adapter} module to incorporate new planners for evaluations. We provide an interface through the abstract base class
\texttt{BaseModelAdapter}, which has four core functions to enable simulation on BridgeSim:
\texttt{load\_model()} for checkpoint initialization and device mapping;
\texttt{prepare\_input()} for transforming raw multi-camera images and ego-state
vectors into the model's native input format;
\texttt{run\_inference()} for executing the forward pass; and
\texttt{parse\_output()} for extracting a predicted waypoint trajectory of shape
$N \times 2$ in the ego-vehicle coordinate frame. Once the model adapter is implemented, it is registered by adding a single entry
to the \texttt{create\_model\_adapter()} factory, after which the new model is
immediately ready for evaluations across all loaded maps and traffic modes.

\textbf{Incorporating New Traffic Modes.}
Traffic behavior is governed by the \texttt{EnvironmentManager} component, which maps symbolic traffic-mode identifiers to their corresponding MetaDrive simulation configurations. To introduce a new traffic mode, such as a learning-based traffic generation model or an adversarial agent policy, contributors simply define a new mode identifier, extend the internal configuration resolver to emit the necessary MetaDrive flags, and optionally provide a custom agent policy class.

\section{Implementation Details on TTA Framework}
\subsection{Implementation Details on Observational Calibrator}
\label{appendix:observational_calibrator}

\begin{figure}[!htbp]
  \centering
  \includegraphics[width=0.95\textwidth]{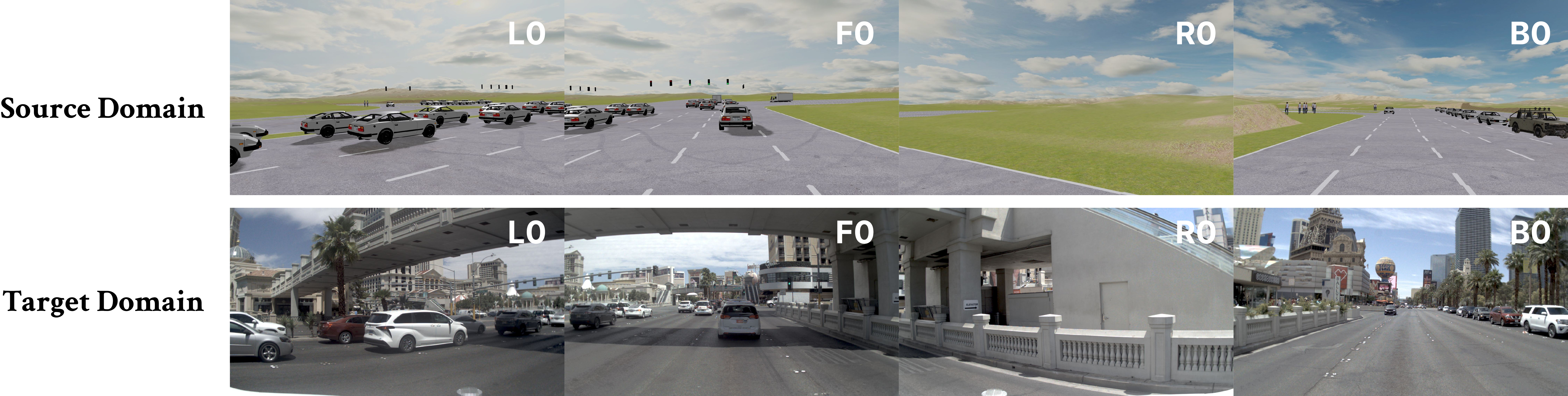}
  \vspace{-1em}
  \caption{Visualization of observation pairs in source and target domains.}
  \label{fig:image_pairs}
\end{figure}

The observational calibrator is designed to align latent representation distributions $\mathcal{D}_{\mathrm{source}}$ and $\mathcal{D}_{\mathrm{target}}$ via a flow-matching framework. To construct the necessary feature pairs for training, we utilize the NAVSIM dataset \cite{navsimv1, navsimv2}, replaying and rendering all annotated frames to generate paired image observations between the source and target domains, as shown in Fig.~\ref{fig:image_pairs}. To process these input observations, we first employ a small encoder that transforms raw scene features into a regularized latent space $\mathcal{Z}$. We then parameterize the vector field $v_\theta(z, t)$ operating on this space using a Diffusion Transformer (DiT)~\cite{Peebles2022DiT} architecture. Specifically, we employ the \textit{DiT-L} configuration with a patch size of 2 to process the latent features, ensuring high-resolution alignment across the domain shift.

Following~\cite{liu2025flowing}, we optimize the calibrator using a multi-objective loss function. In addition to the standard flow-matching objective, the intermediate latent space $\mathcal{Z}$ is explicitly regularized via an \textbf{encoder loss} and a \textbf{KL divergence loss} to preserve semantic consistency and maintain a well-behaved manifold. The total training objective is:
\begin{equation}
    \mathcal{L}_{\mathrm{total}} = \mathcal{L}_{\mathrm{FM}} + \lambda_{\mathrm{enc}} \mathcal{L}_{\mathrm{enc}} + \lambda_{\mathrm{KL}} \mathcal{L}_{\mathrm{KL}},
\end{equation}
where each component is defined as follows:

\begin{itemize}
    \item \textbf{Flow-Matching Loss ($\mathcal{L}_{\mathrm{FM}}$):} We adopt a linear probability path to learn the transport map between source and target distributions. The loss is defined as:
    \begin{equation}
        \mathcal{L}_{\mathrm{FM}}(\theta) = \mathbb{E}_{t, q(z_t|z_0, z_1)} \left[ \| v_\theta(z_t, t) - (z_1 - z_0) \|^2 \right],
    \end{equation}
    where $z_0 \sim \mathcal{D}_{\mathrm{source}}$ and $z_1 \sim \mathcal{D}_{\mathrm{target}}$.
    
    \item \textbf{Encoder Loss ($\mathcal{L}_{\mathrm{enc}}$):} We employ a categorical contrastive loss between source and target feature pairs to regularize the small encoder. This term ensures that latent features corresponding to similar underlying states are clustered together while pushing dissimilar states apart. This prevents mode collapse during the transport process and preserves the discriminative structure of the latent space necessary for the downstream policy $\pi_\phi$.
    
    \item \textbf{KL Loss ($\mathcal{L}_{\mathrm{KL}}$):} To maintain a structured and smooth latent manifold, we apply a Kullback-Leibler divergence penalty against a standard normal prior $\mathcal{P}(z) = \mathcal{N}(0, \mathbf{I})$:
    \begin{equation}
        \mathcal{L}_{\mathrm{KL}} = D_{\mathrm{KL}}(q(z|o) \,\|\, \mathcal{P}(z)).
    \end{equation}
    This variational regularization ensures that the flow-matching trajectory remains within a well-behaved region of the latent space $\mathcal{Z}$.
\end{itemize}

During training, we use the AdamW~\cite{kingma2014adam} optimizer with a learning rate of $1 \times 10^{-4}$ and weight decay of $0.01$. At test time, the calibrated representations are obtained by integrating the learned ODE from $t=0$ to $t=1$ using a first-order Euler solver.

\subsection{More Implementation Details on Test-time Policy Adaptation}
\label{appendix:test-time-adaptation} 

To operationalize the truncated action-value estimator defined in Section~\ref{subsubsec:intrinsic_q_estimation}, we employ Monte Carlo sampling to generate and evaluate future candidate rollouts over the horizon $H$. The empirical reward function $R_k(s_t, \mathbf{a}_t)$ evaluates each candidate prefix by computing \textbf{EPDMS}. Assuming the subscore definitions established in Appendix~\ref{appendix:eval_metrics}, we detail the computational mechanisms enabling their exact and estimated evaluation during real-time closed-loop execution. 

The prefix reward is mathematically formulated as a gated sum of the continuous Quality Features ($\mathcal{F}$), strictly conditioned on the satisfaction of the Critical Constraints ($\mathcal{C}$):
\begin{equation}
    R_k(s_t, \mathbf{a}_t) = \mathbb{I}_{\mathcal{C}} \cdot \sum_{f \in \mathcal{F}} w_f f(s_{t:t+k}, \mathbf{a}_{t:t+k})
\end{equation}
where $\mathbb{I}_{\mathcal{C}} \in \{0, 1\}$ acts as a hard safety filter. If a candidate trajectory violates any critical constraint, $\mathbb{I}_{\mathcal{C}} = 0$, immediately zeroing out the trajectory's expected value.

\textbf{Kinematic Comfort.} Because the test-time adaptation mechanism operates directly on spatial-temporal plans, we have explicit access to both the historical states and the predicted future waypoints of the ego vehicle. Consequently, we do not need to rely on transition approximations; the exact comfort score is computed deterministically by extracting the high-frequency kinematic derivatives (acceleration, jerk, and yaw rate) directly from the sequence of waypoints.

\textbf{Semantic and Map Awareness.} Evaluating map-dependent metrics, such as Drivable Area Compliance (\textbf{DAC}), Driving Direction Compliance (\textbf{DDC}), Traffic Light Compliance (\textbf{TLC}), and Lane Keeping (\textbf{LK}), requires grounding the generated waypoints in the environment's topology. To achieve this, we assume access to an accurate localization model and a robust mapping model, such as High-Definition (HD) Map or other online map generative models. The candidate ego-polygon $\mathcal{P}_{ego}$ is projected at each timestep and geometrically queried against the generated real-time semantic layouts to compute both constraint satisfaction and routing efficiency.

\textbf{Dynamic Safety Estimation.} Evaluating interaction safety, specifically No At-Fault Collisions (\textbf{NC}) and Time-to-Collision (\textbf{TTC}), requires predicting the stochastic futures of other actors. We assume access to the current estimated velocity and acceleration profiles of surrounding agents. Using these profiles, we kinematically propagate their bounding boxes $\mathcal{P}_{obj}$ over the planning horizon to check for intersections with the ego trajectory and to compute the time-to-collision margins.

\textbf{Two-Dimensional Surrogate Safety Measures.} To rigorously evaluate the spatial-temporal proximity to potential collisions across complex driving environments, we utilize a two-dimensional Surrogate Safety Measure (SSM) framework. The foundational metric in this domain is the Time-to-Collision (TTC), which evaluates the temporal proximity to a conflict based on the projected Distance-to-Collision ($DTC$). Operating under a classical assumption of constant velocity at the instantaneous moment of evaluation~\cite{tarko2018surrogate}, the base 2D-TTC is defined using the relative velocity vector $\boldsymbol{v}_{ij}$ between the host vehicle $i$ and an interacting agent $j$:
\begin{equation}
    TTC = \frac{DTC}{\Vert\boldsymbol{v}_{ij}\Vert}
\end{equation}

While computationally straightforward, the standard constant-velocity assumption exhibits significant drawbacks in dynamic, real-world driving scenarios. Specifically, it fails to capture slow, conscious interactive behaviors, such as yielding, braking, or evasive maneuvering, often leading to inaccurate risk assessments during higher-order kinematic interactions~\cite{guo2023modeling, laureshyn2010evaluation}.

To address these limitations, we adopt the Modified Time-To-Collision (MTTC) as our primary safety evaluation metric. The MTTC relaxes the constant-velocity constraint by incorporating the relative acceleration vector, $(\boldsymbol{a}_i - \boldsymbol{a}_j)$, thereby yielding a higher-fidelity, second-order temporal projection. By solving the kinematic displacement equation, the MTTC is formulated as:
\begin{equation}
    MTTC = \frac{-\Vert\boldsymbol{v}_{ij}\Vert \pm \sqrt{\Vert\boldsymbol{v}_{ij}\Vert^2 + 2(\boldsymbol{a}_i - \boldsymbol{a}_j)DTC}}{\boldsymbol{a}_i - \boldsymbol{a}_j}
\end{equation}
Consistent with established collision warning logic, the final MTTC is determined by extracting the minimum positive real root from this quadratic formulation. Furthermore, by structuring the vehicle states into matrices, we enable the parallel computation of the MTTC across all $i,j$ pairs simultaneously. The resulting indicator sets are subsequently filtered according to the behavioral thresholds established by Ozbay et al.~\cite{ozbay2008derivation}, ensuring both high-fidelity risk assessment and computational efficiency for large-scale, multi-agent evaluation.

\section{More Experimental Details}

\subsection{Closed-loop Evaluation Setting}
\label{appendix:cl_eval_details}
\textbf{Scenario Descriptions.} We convert three real-world datasets into BridgeSim platform. \textbf{BridgeSim-NavHard} is derived from the NAVSIM benchmark~\cite{navsimv1}, whose logs originate from the nuPlan dataset~\cite{caesar2021nuplan} at 2~Hz. Agent trajectories are up-sampled to 10~Hz via linear interpolation, and HD maps are extracted from the nuPlan map API with a large spatial buffer to cover the full ego trajectory. The converted scenes contain 421 scenarios. \textbf{BridgeSim-WOMD} is converted from the Waymo Open Motion Dataset~\cite{ettinger2021large}, which provides 9-second scenarios at 10~Hz in a world-frame coordinate system. All agent and map positions are re-centered at the ego vehicle's initial position, and ego velocity is rotated into the vehicle's local frame. The converted scenes contain 400 scenarios. \textbf{BridgeSim-nuScenes} is built on the nuScenes dataset~\cite{caesar2020nuscenes}, which runs at 2~Hz. The converted scenes contain 143 scenarios. Camera images are re-rendered in MetaDrive at 10~Hz, sampled at every fifth step to match the original keyframe rate.

\noindent \textbf{Simulation Settings.} All models are evaluated under simulation operating at a default timestep (\texttt{sim\_dt=0.1s}). Inference is executed at a specified \texttt{replan\_rate}, with cached waypoints consumed between replan intervals. An initial replay phase is performed before evaluation begins to provide historical temporal buffer for agents. For multi-proposal models, a trajectory scorer selects the final plan within a set of trajectory candidates. For running TTA module, an observational calibration module is applied to reduce the domain gap between simulation and real-world training data via a flow matching network with a fixed number of sampling steps.

\subsection{Details on Empirical Study in Section~\ref{subsec:observational_shift}}
\label{appendix:observational_shift_details}

In Fig.~\ref{fig:empirical_observational_shift} from Section~\ref{subsec:observational_shift}, the Average Displacement Error (ADE) comparisons are obtained by evaluating the policy's predicted trajectories over a $4\text{-second}$ horizon against the ground-truth expert demonstrations. To quantify the prediction accuracy accounting for multi-modal outputs, we compute the Minimum Average Displacement Error ($\text{minADE}$) by evaluating the closest predicted trajectory candidate to the ground truth. Let $T$ denote the prediction horizon, $K$ the number of predicted trajectory candidates, $\mathbf{p}_t$ the ground truth spatial coordinates at time step $t$, and $\hat{\mathbf{p}}_t^{(k)}$ the predicted coordinates for the $k$-th candidate. The metric is formulated as:
\begin{equation}
\small
    \text{minADE} = \min_{k \in \{1, \dots, K\}} \frac{1}{T} \sum_{t=1}^{T} \left\| \hat{\mathbf{p}}_t^{(k)} - \mathbf{p}_t \right\|_2,
\end{equation}
where $\| \cdot \|_2$ denotes the Euclidean ($L_2$) norm. Furthermore, the closed-loop performance metrics are evaluated using the \textit{BridgeSim Driving Score} across the BridgeSim-NavHard scenarios.

\subsection{Details on Empirical Study in Section~\ref{subsec:objective_mismatch}}
\label{appendix:objective_mismatch_details}

In the analysis presented in Fig.~\ref{fig:empirical_objective_mismatch}, we employ the \textit{BridgeSim Driving Score} as the empirical realization of the return objective $J(\pi)$. To ensure a rigorous assessment, we evaluate a representative family of state-of-the-art E2E policies~\cite{transfuser, diffusiondrive, diffusiondrivev2, RAP, DrivoR}, reporting the upper and lower performance bounds to characterize the variance across different architectural paradigms. The experimental protocols are defined as follows:
\begin{itemize}
    \item Fig.~\ref{fig:empirical_objective_mismatch}(a): we evaluate policies over a fixed simulation horizon of $T=80$ in BridgeSim-NavHard. To isolate the impact of the execution horizon $k$, we compare a baseline of fixed $k$-step prefixes against an ``optimal execution'' baseline derived via exhaustive search over the policy's predicted trajectory set.
    \item Fig.~\ref{fig:empirical_objective_mismatch}(b): we contrast the performance of policies operating under two distinct selection criteria in BridgeSim-Navhard: 1) the \textbf{OL Objective}, which relies on the agent's learned Q-value estimator, and 2) the \textbf{CL Objective}, where actions are selected via the ground-truth EPDMS metric to simulate an unbiased value function. The evaluation is done across ($T \in \{40, 80, 120, 160\}$).
    \item Fig.~\ref{fig:empirical_objective_mismatch}(c): we compute the Pearson correlation between open-loop and closed-loop metrics across the policies. The OL metric is obtained through non-reactive simulation ($T=40$) following the NAVSIMV2 protocol~\cite{navsimv2} in BridgeSim-NavHard. In contrast, the CL metric is derived from the BridgeSim Driving Score across expanding simulation horizons ($T \in \{40, 80, 120, 160\}$) to observe the decay of predictive validity over time in BridgeSim-NavHard.
\end{itemize}

\begin{figure}[tbh]
  \centering
  \includegraphics[width=0.95\textwidth]{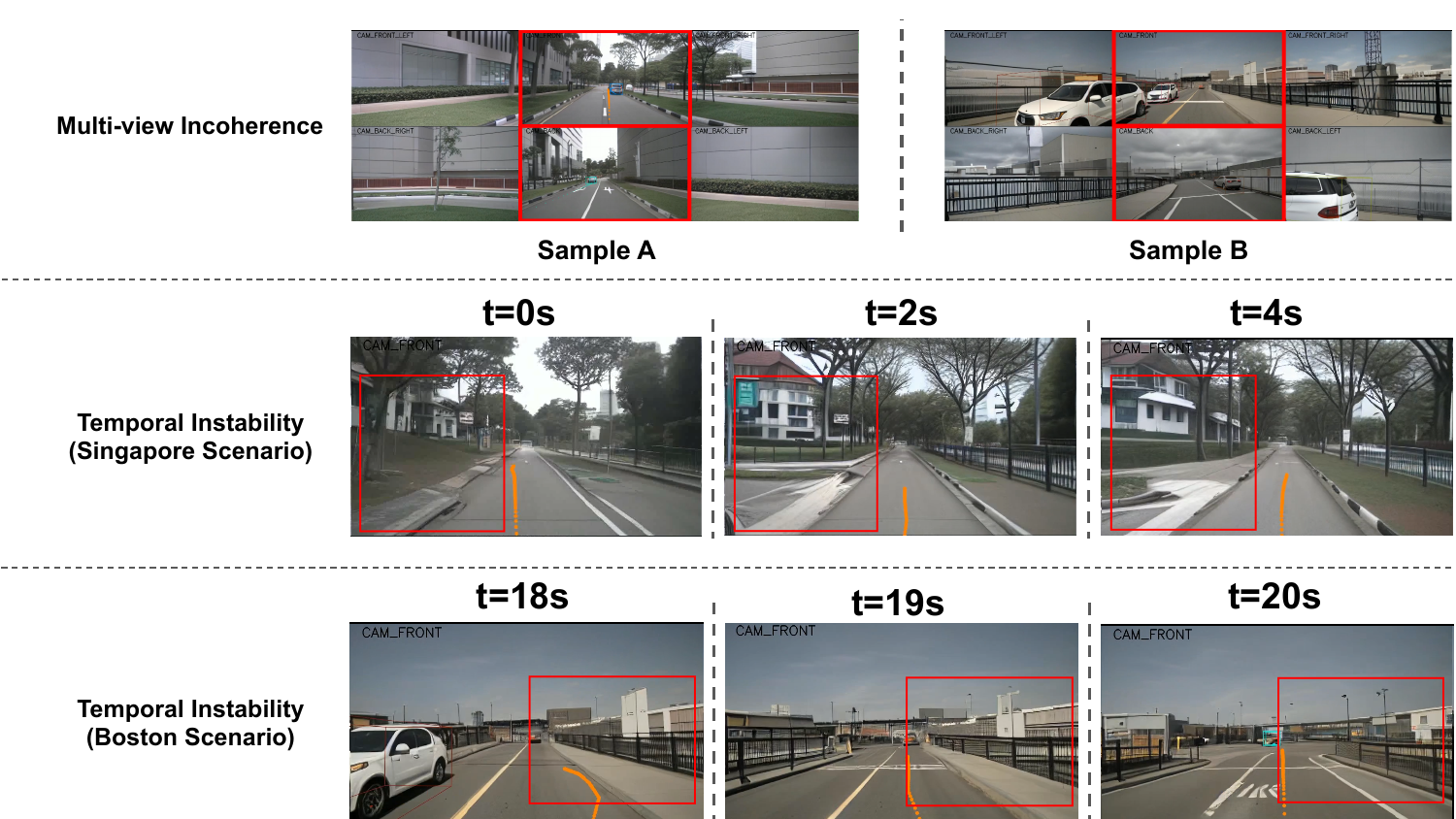}
  \vspace{-1em}
  \caption{Qualitative analysis of simulation artifacts in DriveArena: multi-view incoherence (left) and temporal instability (right). Red boxes highlight inconsistencies that lead to policy failures or metric distortion.}
  \label{fig:drivearena_artifacts}
\end{figure}

\begin{figure}[tbh]
  \centering
  \includegraphics[width=0.95\textwidth]{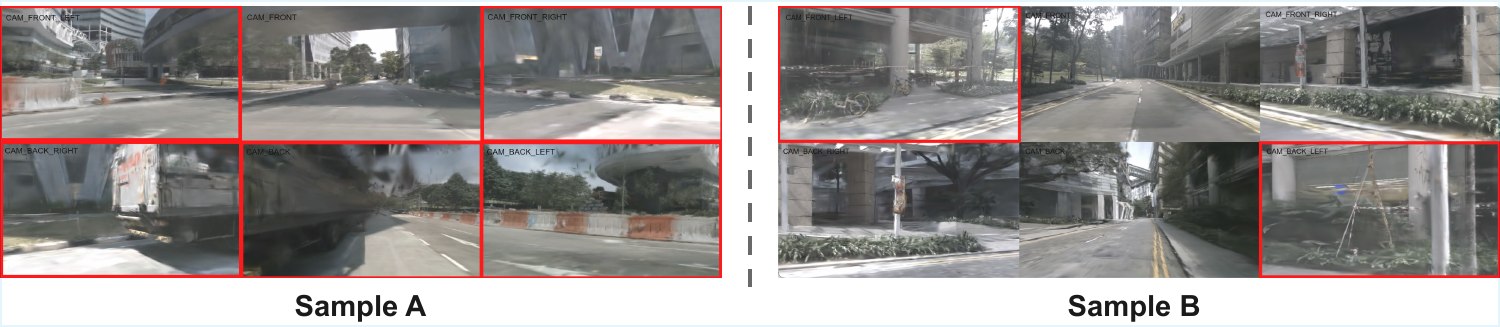}
  \vspace{-1em}
  \caption{Qualitative analysis of reconstruction artifacts and ghosting effects in HUGSIM. Red boxes highlight inconsistencies. }
  \label{fig:hugsim_artifacts}
\end{figure}

\begin{table}[tbh]
\centering
\caption{Performance comparisons with different reactive traffic modes with complete sub-metric scores. }
\label{tab:complete_reactive_cl_scores}
\renewcommand{\arraystretch}{1.5} 
\setlength{\aboverulesep}{0pt}    
\setlength{\belowrulesep}{0pt}    
\resizebox{\textwidth}{!}{ 
\begin{tabular}{l *{11}{c} @{\hspace{6pt} } *{11}{c}}
\toprule
\multirow{2}{*}{\textbf{Model}} & \multicolumn{11}{c}{\textbf{IDM}} & \multicolumn{11}{c}{\textbf{Safety-Critical}} \\
\cmidrule(lr){2-12} \cmidrule(lr){13-23}
& \textbf{DS} & \textbf{EPDMS} & \textbf{RC} & \textbf{NC} & \textbf{DAC} & \textbf{DDC} & \textbf{TL} & \textbf{TTC} & \textbf{LK} & \textbf{HC} & \textbf{EC} & \textbf{DS} & \textbf{EPDM} & \textbf{RC} & \textbf{NC} & \textbf{DAC} & \textbf{DDC} & \textbf{TL} & \textbf{TTC} & \textbf{LK} & \textbf{HC} & \textbf{EC} \\
\midrule
DiffusionDrive & 45.30 & 72.95 & 59.93 & 95.41 & 85.47 & \textbf{100.00} & 98.36 & 89.73 & 89.83 & 92.31 & 40.82 & 38.68 & 68.16 & 54.77 & 93.62 & 83.86 & \textbf{100.00} & 98.36 & 84.97 & 90.71 & 89.56 & 39.13 \\
DiffusionDriveV2 & 31.94 & 65.43 & 45.85 & 94.57 & 82.09 & \textbf{100.00} & 98.23 & 89.10 & 85.42 & 74.52 & 22.99 & 28.17 & 61.45 & 43.50 & 93.15 & 80.78 & \textbf{100.00} & 98.30 & 85.72 & 85.81 & 70.99 & 21.68 \\
RAP & 60.02 & 76.85 & 77.20 & 97.68 & 88.84 & \textbf{100.00} & 98.49 & 93.07 & 78.16 & \textbf{94.49} & \textbf{53.84} & 51.83 & 70.88 & 71.22 & 95.03 & 88.16 & \textbf{100.00} & 98.58 & 86.37 & 78.26 & \textbf{90.22} & \textbf{50.53} \\
\midrule
\rowcolor{green!15} TTA \scriptsize (DiffusionDrive) & 62.68 \scriptsize \textcolor{red}{+17.4} & 84.36 & 72.28 & 96.85 & 95.63 & \textbf{100.00} & 98.98 & 91.57 & \textbf{92.43} & 87.95 & 40.13 & 55.02 \scriptsize \textcolor{red}{+16.3} & 78.95 & 67.92 & 94.73 & 95.21 & \textbf{100.00} & 98.76 & 86.64 & \textbf{91.30} & 83.65 & 37.11 \\
\rowcolor{green!15} TTA \scriptsize (DiffusionDriveV2) & 53.96 \scriptsize \textcolor{red}{+22.0} & 77.47 & 67.89 & 96.08 & 90.53 & \textbf{100.00} & \textbf{99.43} & 89.46 & 89.98 & 85.46 & 29.90 & 47.64 \scriptsize \textcolor{red}{+19.5} & 71.66 & 65.91 & 93.54 & 89.16 & \textbf{100.00} & 99.07 & 84.23 & 90.11 & 81.73 & 27.49 \\
\rowcolor{green!15} TTA \scriptsize (RAP) & \textbf{79.07} \scriptsize \textcolor{red}{+19.1} & \textbf{90.92} & \textbf{86.21} & \textbf{98.28} & \textbf{98.62} & \textbf{100.00} & 99.02 & \textbf{95.31} & 91.52 & 92.60 & 51.38 & \textbf{65.94} \scriptsize \textcolor{red}{+14.1} & \textbf{82.01} & \textbf{78.66} & \textbf{95.30} & \textbf{96.19} & \textbf{100.00} & \textbf{99.14} & \textbf{87.71} & 90.26 & 88.18 & 48.07 \\
\bottomrule
\end{tabular}
}
\end{table}

\subsection{Details on Experimental Results in Reactive Evaluation}
\label{appendix:detailed_reactive_metrics}
We show the complete sub-metric scores in reactive evaluations results (see Table~\ref{tab:reactive_cl_scores}) in Table~\ref{tab:complete_reactive_cl_scores}.

\subsection{More Discussions of OL Training and Evaluations}
\label{subsec:discussions}
\textbf{Will RL finetuning on OL proxy reward benefit CL performance?} Recent papers~\cite{diffusiondrivev2, zhou2025autovla} advocate for RL finetuning to enhance planning performance, primarily utilizing open-loop proxy rewards as the training objective. However, as shown in Table~\ref{tab:reactive_cl_scores}, our comparison between DiffusionDrive and DiffusionDriveV2 reveals a significant discrepancy between open-loop and closed-loop outcomes. Specifically, DiffusionDriveV2 exhibits a pronounced performance degradation in CL simulations while it obtains higher OL metrics in NAVSIM. This suggests that optimizing the OL objective doesn't guarantee the optimization of the CL objective.

\textbf{To what extent can we rely on OL metrics to predict CL performance?} Fig.~\ref{fig:empirical_objective_mismatch}(c) indicates that while OL metrics maintain predictive validity for short horizons ($T=H$), this correlation decays rapidly as $T$ exceeds $H$. This decaying trend indicates that while high OL scores signify that a policy possesses sufficient local planning proficiency and safety for in-distribution scenarios, they do not serve as a reliable predictor of CL performance in reactive or out-of-distribution environments. These findings suggest two critical takeaways: 1) scaling open-loop training data doesn't statistically guarantee improvements for closed-loop performance, and 2) effective policies must prioritize objectives that approximate gold Q-value estimations to ensure long-horizon stability. 

\section{Limitations of Existing E2E Closed-loop Benchmarks}
\label{appendix:current_e2e_bottlenecks}

Existing end-to-end closed-loop benchmarks, such as DriveArena~\cite{yang2024drivearena} and HUGSIM~\cite{zhou2024hugsim}, exhibit limitations stemming from \textit{temporal and semantic inconsistencies} during simulation.

\textbf{Generative Artifacts in DriveArena.} 
As illustrated in Fig.~\ref{fig:drivearena_artifacts}, DriveArena displays notable simulation discrepancies. \textit{1) Multi-view Incoherence:} in the multi-view comparison, discrepancies in the color, length, and placement of lane markings between front and rear camera views (red boxes) can lead to oscillatory policy behavior due to contradictory observations. \textit{2) Temporal Instability:} as shown in the temporal sequences, road features such as intersections are inconsistently rendered or spontaneously modified across sequential timestamps (e.g., $t=2s$ in Singapore and $t=19s$ in Boston, marked by red boxes), resulting in unstable scene geometry.

\textbf{Semantic Mismatch in HUGSIM.} 
As shown in Fig.~\ref{fig:hugsim_artifacts}, 3D Gaussian Splatting-based methods like HUGSIM encounter challenges with dynamic actor integration. The Sample A \& B highlight simulation artifacts and blurring around moving vehicles (red boxes), which are frequently misinterpreted as obstacles. Furthermore, the misalignment between visual rendering and the underlying HD map creates semantic inconsistency.

\textbf{Structural Invariance in BridgeSim.} 
In contrast, BridgeSim maintains strict structural invariance by decoupling vehicle dynamics from visual rendering. By leveraging the MetaDrive physics engine, the kinematic state $s_t$ serves as the deterministic reference for the environment. Sensor observations $o_t$ are generated via geometric projection rather than generative synthesis, ensuring spatial and temporal consistency across all camera frames and timestamps.

\section{Future Work}
We expect that future research will focus on building more realistic evaluations and developing methods that enhance driving agents' safety and robustness under real-world constraints:

\textbf{1. E2E Closed-loop Benchmarks.} Existing CL benchmarks remain limited by observational fidelity, temporal inconsistencies, and evaluation efficiency (see Appendix~\ref{appendix:current_e2e_bottlenecks}). Future benchmarks should move toward unified, hybrid-engine frameworks that combine the controllability of game-engine simulation with the diversity of data-driven world models, enabling scalable yet realistic stress testing of long-horizon CL simulation.

\textbf{2. E2E Closed-loop Training.} Current OL pretraining methods lack error-recovery samples and feedback-driven learning objectives to teach a policy to safely navigate dynamic and reactive environments. Advancing CL training methods, such as scalable variants of RoAD~\cite{garcia2025road}, learning with world modeling, and reinforcement learning with reliable reward formulation, remains critical. In particular, defining and learning unbiased value functions that reflect true CL returns in real-world driving is a key open challenge.

\end{document}